# Cause Identification from Aviation Safety Incident Reports via Weakly Supervised Semantic Lexicon Construction

**Muhammad Arshad Ul Abedin**               ARSHAD@STUDENT.UTDALLAS.EDU
**Vincent Ng**                                   VINCE@HLT.UTDALLAS.EDU
**Latifur Khan**                                 LKHAN@UTDALLAS.EDU
*Department of Computer Science*
*Erik Jonsson School of Engineering & Computer Science*
*The University of Texas at Dallas*
*800 W. Campbell Road; MS EC31*
*Richardson, TX 75080 U.S.A.*

## Abstract

The Aviation Safety Reporting System collects voluntarily submitted reports on aviation safety incidents to facilitate research work aiming to reduce such incidents. To effectively reduce these incidents, it is vital to accurately identify *why* these incidents occurred. More precisely, given a set of possible causes, or *shaping factors*, this task of cause identification involves identifying all and only those shaping factors that are responsible for the incidents described in a report. We investigate two approaches to cause identification. Both approaches exploit information provided by a semantic lexicon, which is automatically constructed via Thelen and Riloff's Basilisk framework augmented with our linguistic and algorithmic modifications. The first approach labels a report using a simple heuristic, which looks for the words and phrases acquired during the semantic lexicon learning process in the report. The second approach recasts cause identification as a text classification problem, employing supervised and transductive text classification algorithms to learn models from incident reports labeled with shaping factors and using the models to label unseen reports. Our experiments show that both the heuristic-based approach and the learning-based approach (when given sufficient training data) outperform the baseline system significantly.

## 1. Introduction

Safety is of paramount importance when it comes to the aviation industry. In 2007 alone, there were 4659 incidents[1], including 26 fatal accidents with 750 casualties[2]. To improve the aviation safety situation, the Aviation Safety Reporting System (ASRS) was established in 1976 to make safety incident data available to researchers. ASRS collects voluntarily submitted reports about aviation safety incidents written by flight crews, attendants, controllers and other related parties. The reports contain a number of *fixed fields* and a *free text narrative* describing the incident. However, the data has grown to be quite large over the years and it is getting increasingly difficult, if not impossible, to analyze these reports by human means. It has become necessary that these reports be analyzed through automated means.

---

1. http://asrs.arc.nasa.gov/
2. http://www.flightsafety.gov/

To take full advantage of this data to reduce safety incidents, it is necessary to extract from the reports both *what* happened and *why*. Once both are known, then it is possible to identify the correlations between the incidents and their causes, and take fruitful measures toward eliminating the causes. However, the fixed fields in the reports are devoted to various aspects of *what* happened during the incidents, and there is no fixed field that indicates the incidents' causes. Instead, the reporter discusses in the report narrative what he thinks caused the incident, along with the incident description. Thus the cause of the incident has to be extracted by analyzing the free text narrative. As an example, a report is shown next to illustrate the task:

> **Report#424362.** WHILE descending into lit we encountered Instrument Meteorological Conditions; rime ice; rain; and moderate chop. as I turned to a heading with the Auto-Pilot direct lit the attitude indicator remained in a bank. XCHKING; I noticed the Radio Magnetic Indicator'S were 55 degree off headings. I switched to #2 and corrected the course. the Auto-Pilot and flight director were kicked off. I continued to have problems with the altitude select and Auto-Pilot as I attempted to re-engage it. it was during radar vectors to the approach and descent to 2300 feet that we noticed our altitude at 2000 feet Mean Sea Level. we stopped the descent and climbed to 2300 feet Mean Sea Level. Air Traffic Control noted our altitude deviation at the time we noticed. we were thankful for their backup during a time of flight director problems in our cockpit. this occurred at the end of a 13 hour crew day; bad weather; instrument problems; and lack of crew rest. the First Officer (Pilot Not Flying) in the right seat; had only 4 hours of rest due to inability to go to sleep the night before. we were tired from a trip lit-ORL-lit. we had not eaten in about 7 hours.[3]

Posse et al. (2005) identify 14 most important cause types, or *shaping factors*, that can influence the occurrence of the aviation safety incident described in an ASRS report. These shaping factors are the contextual factors that influenced the reporter's behavior in the incident and thus contributed to the occurrence of the incident. Some of these factors can be attributed to humans (e.g., a pilot or a flight attendant has psychological *Pressure*, an overly heavy *Taskload*, or an unprofessional *Attitude* that impacts his performance), while some are related to the surrounding environment (e.g., *Physical Environment* such as snow, and *Communication Environment* such as auditory interference). A detailed description of these 14 shaping factors can be found in Section 2.1.

In the above report, we find that the incident was influenced by three shaping factors, namely *Physical Environment* (which concerns bad weather, as mentioned above), *Resource Deficiency* (which concerns problems with the equipment), and *Duty Cycle* (which refers to physical exhaustion due to long hours of duty without adequate rest or replenishment). These three shaping factors are indicated by different words and phrases in the report. For instance, the bad weather condition is expressed using phrases such as *rime ice*, *rain* and *moderate chop*, while the details of the equipment problem appear as sentence fragments like

---

3. To improve readability, the report has been preprocessed from its original form using the steps described in Section 2.2.

*attitude indicator remained in a bank*, *55 degree off headings* and *flight director problems*. The issue with the long hours of duty is illustrated by the sentence fragments like *13 hour crew day* and *tired from a trip*. The goal of our cause identification task for the aviation safety domain, then, is to identify which of the 14 shaping factors contributed to the incident described in a report using the lexical cues appearing in the report narrative.

However, as mentioned earlier, the sheer volume of the data makes it prohibitive to analyze all the reports manually and identify the associated shaping factors. Thus, the focus of our research is *automated cause identification* from the ASRS reports, which involves automatically analyzing the report narrative and identifying the responsible shaping factors. This brings our problem into the domain of Natural Language Processing (NLP).

Since we have a set of texts (i.e., the report narratives) and a set of possible labels for these texts (i.e., the shaping factors), this task is most naturally cast as a text classification task. However, unlike topic-based text classification, cause-based text classification has not been addressed extensively in the NLP community. Previous work on causal analysis is quite different in nature from our cause-based text classification task. More specifically, previous cause analysis works do not involve text classification, focusing instead on determining the existence of a *causal relation* between two sentences or events. For instance, there has been some work on causal analysis for question answering, where a question may involve the cause(s) of an event (e.g., Kaplan & Berry-Rogghe, 1991; Garcia, 1997; Khoo, Chan, & Niu, 2000; Girju, 2003). Here, the focus is on finding *causal relationship between two sentence components*. As another example, causal analysis on equipment malfunction reports have been attempted by Grishman and Ksiezyk (1990), whose work is restricted to the analysis of reports related to one specific piece of equipment they studied. They analyze *cause-effect relations between events* leading to the malfunction described in the reports.

Cause identification from aviation safety reports is a rather challenging problem, as a result of a number of factors specific to the ASRS dataset. First, unlike many NLP problems where the underlying corpus is composed of a set of well-edited texts such as newspaper reports, reviews, legal and medical documents[4], the ASRS reports are mostly written in informal manner, and since they have not been edited except for removing author-identity information, the reports tend to contain spelling and grammatical mistakes. Second, they employ a large amount of domain-specific acronyms, abbreviations and terminology. Third, the incident described in a report may have been caused by more than one shaping factor. Thus reports can have multiple shaping factor labels, making the task more challenging than binary classification, or even multi-class problems where each instance has only one label. Above all, the scarcity of labeled data for this task, coupled with highly imbalanced class distributions, makes it difficult to acquire an accurate classifier via supervised learning.

Previous work on cause identification for the ASRS reports was done primarily by the researchers at NASA (see Posse et al., 2005) and, to our knowledge, has involved manual analysis of the reports. Specifically, NASA brought together experts on aviation safety, human factors, linguistics and English language to participate in a series of brainstorming sessions, and generated a collection of seed keywords, simple expressions and template expressions related to each shaping factor. Then they labeled the reports with the shaping factors by looking for the related expressions in the report narrative. However, there is a major weakness associated with this approach: it involves a large amount of human effort on identifying the relevant keywords and expressions, and yet the resulting list of keywords and expressions is by no means exhaustive. Moreover, they evaluated their approach on only 20 manually labeled reports. Such a small-scale evaluation is by no means satisfactory as judged by current standard in NLP research. One of our contributions in this research is the annotation of 1333 ASRS reports with shaping factors, which serve as a standard evaluation dataset against which different cause identification methods can be compared.

---

4. Recently, work has started on processing *blogs*, which may not be so grammatical either, but blogs typically are not full of domain-specific terminology.

In this paper, we investigate two alternative approaches to cause identification, both of which exploit information provided by an automatically constructed semantic lexicon. More specifically, in view of the large amount of human involvement in NASA's work, we aim to replace the manual selection of seed words with a bootstrapping approach that automatically constructs a *semantic lexicon*. Specifically, motivated by Thelen and Riloff's (2002) Basilisk framework, we learn a semantic lexicon, which consists of a set of words and phrases semantically related to each of the shaping factors, as follows. Starting from a small set of *seed* words and phrases, we augment these seeds in each iteration by automatically finding a fixed number of words and phrases related to the seeds from the corpus and adding them to the seed list. Most importantly, however, we propose four modifications to the Basilisk framework that can potentially improve the quality of the generated lexicon. The first is a linguistic modification: in addition to using parse-based features (e.g., subject-verb and verb-object features) as in Basilisk, we employ features that can be computed more robustly (e.g., N-grams). The remaining three are all algorithmic modifications to the Basilisk framework, involving (1) the use of a probabilistic semantic similarity measure, (2) the use of a common word pool, and (3) the enforcement of minimum support and maximum generality constraints for words and their extraction patterns, which favors the addition of frequently-occurring content-bearing words and disfavors overly-general extraction patterns.

As mentioned above, we investigate two approaches to cause identification that exploit the automatically learned semantic lexicon. The first approach is a *heuristic* approach, which, motivated by Posse et al. (2005), labels a report with a shaping factor if it contains at least a word or a phrase that is relevant to the shaping factor. Unlike Posse et al.'s work, where these relevant words and phrases employed by the heuristic procedure are all manually identified, we automatically acquire these words and phrases via the semi-supervised semantic lexicon learning procedure described above. The second approach is a *machine-learning* approach that is somewhat orthogonal to NASA's approach: instead of having a human identify seed words and phrases relevant to each shaping factor, we have humans annotate a small subset of the available incident reports with their shaping factors, and then apply a machine learning algorithm to train a classifier to automatically label an unseen report, using combinations of N-gram features and words and phrases automatically acquired by the aforementioned semantic lexicon learning procedure. As we will see, we acquire this cause identifier using Support Vector Machines (SVMs), which have been shown to be effective for topic-based text classification. Since we only have a small number of labeled reports, we also attempt to combine labeled and unlabeled reports using the transductive version of SVMs.

Since our approaches rely on simple linguistic knowledge sources that involve N-grams and words and phrases automatically acquired during the semantic lexicon learning procedure, one may argue that the use of these simple features are not sufficient for cause identification. It is important to point out that we are by no means arguing that these features are sufficient for cause identification. However, the use of these simple features is *relevant* for the task and is motivated by the work performed by the NASA researchers, who, as mentioned above, have manually identified seed words and phrases for each shaping factor (Posse et al., 2005). Our semantic lexicon learning procedure precisely aims to learn such words and phrases. While our error analysis reveals that these simple linguistic features are not sufficient for learning cause identification (and that more sophisticated knowledge sources are needed to improve performance), as one of the first attempts to tackle this cause identification task, we believe that the use of these simple features is a good starting point and establishes a baseline against which future studies on this domain-specific problem can be compared.

We evaluate the aforementioned two approaches on our manually annotated ASRS reports. Our experiments show a number of interesting results. First, the best performance is achieved using the heuristic approach, where we label a report on the basis of the presence of the automatically acquired lexicon words and phrases in the report, achieving an F-measure of 50.21%. More importantly, this method significantly surpasses the performance of our baseline system, which labels a report on the basis of the presence of a small set of manually identified seed words and phrases. These results suggest that employing an automatically acquired semantic lexicon is relevant and useful for cause-based text classification of the ASRS reports. Second, the words and phrases in the learned semantic lexicon, when used as features for training SVMs in the classification approach, do not improve the performance of an SVM classifier that is trained solely on N-gram based features when the amount of training data is small. However, when we increase the amount of training data (by cross-validation), using the lexicon words and phrases as features in addition to unigrams and bigrams helps improve classifier performance statistically significantly. In particular, we have observed an F-measure of 53.66% from the SVM classifiers using a combination of unigrams, bigrams and lexicon words and phrases as features. These results again confirm that the words and phrases from the learned semantic lexicon are relevant and valuable features for identifying the responsible shaping factors. Nevertheless, the magnitude of the improvement indicates that there is still much room for improvement, which may be achieved by using deeper semantic features.

In summary, we believe that our work on automated cause identification makes five primary contributions:

- We show that instead of manually analyzing all the incident reports to identify the relevant shaping factors, it is possible to reduce the amount of human effort required for this task by manually analyzing only a small subset of the reports and identifying the shaping factors of the rest of the reports by using automated methods.

- We propose several modifications to Thelen and Riloff's (2002) semi-supervised lexicon learning framework, and show that our Modified Basilisk framework allows us to acquire a semantic lexicon that yields significantly better performance for cause identification than the original Basilisk framework. Equally importantly, none of our modifications are geared towards the cause identification task, and hence they are applicable more generally to the semantic lexicon learning task. In fact, our addi-tional experiments suggest that Modified Basilisk yields better accuracy than Original Basilisk when bootstrapping general semantic categories.

- We show that semantic lexicon learning is useful for cause identification from the ASRS reports. In particular, the words and phrases from the learned semantic lexicon can be profitably used to improve both a heuristic-based approach and a learning-based approach (when given sufficient training data) to cause identification. In addition, we believe that in any similar cause identification task where the causes are described in the text, it may be useful to learn a semantic lexicon containing key words and phrases related to the different types of possible causes and use these key words and phrases as features for machine learning.

- In an attempt to deduce the weaknesses of our approaches and help direct future research, we have performed an analysis of the errors made by the best-performing system, namely the heuristic approach using the semantic lexicon learned by our modified Basilisk method on a randomly chosen subset of the test reports.

- We have manually annotated a subset of the reports with the relevant shaping factors. This set of annotated reports, which have been made publicly available, can serve as a standard evaluation set for this task in future research and also for comparing to other approaches to cause identification.

The rest of the paper is organized as follows. In Section 2, we discuss the dataset, the shaping factors, and how the reports were preprocessed and annotated. Section 3 defines the baseline, which simply looks for a small set of manually extracted seed words and phrases in the report narratives. In Section 4, we describe our semantic lexicon learning procedure, which is based on the Basilisk lexicon learning procedure (Thelen & Riloff, 2002) augmented with our modifications. In Section 5, we discuss our heuristic-based and learning-based approaches to cause identification. We evaluate these two approaches in Section 6 and discuss related work in Section 7. Finally, in Section 8, we summarize our conclusions and discuss future work.

## 2. Dataset

The dataset used in this research is the aviation safety incident reports publicly available from the website of Aviation Safety Reporting System[5]. We used all 140,599 reports collected during the period from January 1998 to December 2007. Each report contains a free text narrative written by the reporter and several fixed fields about the incident like the time and place of the incident, environment information, details about the aircrafts involved, the reporting persons' credentials, details like anomaly, detector, resolution and consequence about the incident itself, and a description of the situation. In other words, the fixed fields in a report contain various information about *what* happened, and under what physical circumstances, but do not cover *why* the incident took place. As discussed by Posse et al. (2005) and Ferryman, Posse, Rosenthal, Srivastava, and Statler (2006), only the narrative of a report contains information on the shaping factors of the incident. For

---

this reason, we decided to analyze only the free-text narrative of a report using NLP techniques to identify what the shaping factor(s) of the incident may be, and we constructed the corpus for this task by combining the narratives of these 140,599 reports.

## 2.1 Shaping Factors

The incidents described in the ASRS reports happen for a variety of reasons. Posse et al. (2005) focus on the 14 *shaping factors*, or simply *shapers*. Following is a short description of these shaping factors, taken verbatim from the work of Posse et al..

1. **Attitude:** Any indication of unprofessional or antagonistic attitude by a controller or flight crew member.

2. **Communication Environment:** Interferences with communications in the cockpit such as noise, auditory interference, radio frequency congestion, or language barrier.

3. **Duty Cycle:** A strong indication of an unusual working period e.g., a long day, flying very late at night, exceeding duty time regulations, having short and inadequate rest periods.

4. **Familiarity:** Any indication of a lack of factual knowledge, such as new to or unfamiliar with company, airport, or aircraft.

5. **Illusion:** Illusions include bright lights that cause something to blend in, black hole, white out, or sloping terrain.

6. **Physical Environment:** Unusual physical conditions that could impair flying or make things difficult, such as unusually hot or cold temperatures inside the cockpit, cluttered workspace, visual interference, bad weather, or turbulence.

7. **Physical Factors:** Pilot ailment that could impair flying or make things more difficult, such as being tired, fatigued, drugged, incapacitated, influenced by alcohol, suffering from vertigo, illness, dizziness, hypoxia, nausea, loss of sight, or loss of hearing.

8. **Preoccupation:** A preoccupation, distraction, or division of attention that creates a deficit in performance, such as being preoccupied, busy (doing something else), or distracted.

9. **Pressure:** Psychological pressure, such as feeling intimidated, pressured, pressed for time, or being low on fuel.

10. **Proficiency:** A general deficit in capabilities, such as inexperience, lack of training, not qualified, not current, or lack of proficiency.

11. **Resource Deficiency:** Absence, insufficient number, or poor quality of a resource, such as overworked or unavailable controller, insufficient or out-of-date chart, equipment malfunction, inoperative, deferred, or missing equipment.

12. **Taskload:** Indicators of a heavy workload or many tasks at once, such as short-handed crew.

13. **Unexpected:** Something sudden and surprising that is not expected.

14. **Other:** Anything else that could be a shaper, such as shift change, passenger discomfort, or disorientation.

## 2.2 Preprocessing

For our semantic lexicon learning approach to cause identification, we need to identify (1) the part-of-speech (POS) of each word in the text, (2) the phrases or chunks in the sentences, and (3) the grammatical roles of the words and their governing words. Ideally, to achieve high accuracies on these three tagging tasks, we would manually annotate a section of the ASRS corpus with the appropriate annotations (e.g., POS tags, chunks) and train appropriate taggers on it to tag the rest of the corpus. However, this by itself is a labor-intensive task, and is beyond the scope of this paper. Therefore, we have used publicly available tools trained on standard corpora for these three tasks. It is inevitable that this will not produce the most accurate automatic annotations of our corpus, but as we will see, this has not caused problem in this task.

From our corpus, we first identify sentence boundaries using the tool MXTERMINA-TOR[6]. Second, we run the POS tagger CRFTagger (Phan, 2006b), which uses the Penn Treebank tag set (Marcus, Santorini, & Marcinkiewicz, 1993), on the sentences detected by MXTERMINATOR. Third, we run the chunker CRFChunker (Phan, 2006a) on the tagged text to identify different types of phrases. Also, the Minipar parser (Lin, 1998) is run on the sentences to identify the grammatical roles of the words. However, the report text has to be preprocessed before applying these tools for reasons described in the following paragraphs.

The reports in the ASRS data set are usually informally written, using various domain specific abbreviations and acronyms. In general, as observed by van Delden and Gomez (2004), Posse et al. (2005) and Ferryman et al. (2006), these narratives tend to be written in short, abbreviated manner, and tend to contain poor grammar. Also, the text has been converted to all upper-case. Following is an example of the narrative of a report:

> TAXIING FROM THE RAMP AT LAF AT NIGHT. MADE A WRONG TURN AND CROSSED RWY 10/28; THE ACTIVE AT THE TIME. THERE WAS NO SIGN TO INDICATE WHICH RWY I WAS XING. I CLRED BOTH DIRECTIONS BEFORE XING. WE WERE THE ONLY ACFT ON THE FIELD AT THE TIME. NO MENTION ON THE ATIS OF SIGNS BEING OUT OR CONSTRUCTION ON THE RAMP AREA. THE CTLR DIDN'T QUESTION US; IT WAS I WHO BROUGHT THE SIT UP AFTER I HAD CROSSED THE ACTIVE RWY. COMMUTER OPS OF 3 DAYS OF HVY FLYING; REDUCED REST; NO RWY SIGNS AND BUSY DOING LAST MIN COMMUTER PAPER WORK CHANGES; ALL CONTRIBUTED TO THE RWY INCURSION. 12 HR DAY 6 HR FLT TIME.

---

6. `ftp://ftp.cis.upenn.edu/pub/adwait/jmx/`, trained on the Wall Street Journal corpus

These reports need some preprocessing before NLP techniques can be applied to them, since these off-the-shelf tools (e.g., the POS tagger) were all trained on mixed-case texts. For example, running CRFTagger (which was trained on the WSJ corpus with correct cases) on the first two sentences yield the following:

1. TAXIING/NNP FROM/NNP THE/DT RAMP/NNP AT/IN LAF/NNP AT/IN NIGHT/NN ./.

2. MADE/NNP A/DT WRONG/NNP TURN/NNP AND/CC CROSSED/VBD RWY/NNP 10/28/CD ;/: THE/DT ACTIVE/NNP AT/IN THE/DT TIME/NN ./.

As can be seen, the tagger mislabels the words TAXIING, FROM, MADE, WRONG and ACTIVE as proper nouns (NNP), instead of tagging them as verb, preposition, verb, adjective and adjective respectively. This occurs because a good feature for detecting proper nouns in a sentence is the case of its first character. Since all the words begin with a capital letter, the tagger mistakes a significant portion of these words as NNP. Another reason that the tagger performs poorly on this corpus is that a lot of abbreviations appear in the text. For example, XING and HVY are short for *crossing* and *heavy*. But since they are not likely to be known to a POS tagger trained on a standard well-edited corpus, they would be identified as unknown words, and most likely be tagged as nouns instead of verb and adjective respectively. Similar problems have been observed for the parsers and chunkers. For this reason, we decided to preprocess the text by expanding the abbreviations and restoring the cases of the words.

To expand the acronyms and abbreviations, we rely on the official list of acronyms and abbreviations used in the ASRS reports[7]. In a small number of cases, the same abbreviation or acronym may have more than one expansion. For example, *ARR* may mean either *arrival* or *arrive*. In such cases we arbitrarily chose one of the possibilities[8]. Then, to restore case, a set of English word lists, place names and person names[9] were applied to the text to identify the known words. If a word in the report text appeared in the word lists, then it was converted to lower case. All the other unknown words were left uppercase. The result of this process on the aforementioned narrative is as follows:

TAXIING from the ramp at LAF at night. made a wrong turn and crossed runway 10/28; the active at the time. there was no sign to indicate which runway I was crossing. I cleared both directions before crossing. we were the only aircraft on the field at the time. no mention on the Automatic Terminal Information Service of signs being out or construction on the ramp area. the controller DIDN'T question us; it was I who brought the situation up after I had crossed the active runway. commuter operations of 3 days of heavy flying;

---

7. See `http://akama.arc.nasa.gov/ASRSDBOnline/pdf/ASRS_Decode.pdf`.

8. A better option would be to disambiguate between the alternative expansions based on context (e.g., the method followed by Banko & Brill, 2001). However, the number of such ambiguities in the acronyms and abbreviations list is small (10, to be exact), and they are either the same POS or variations of the same word. Thus the effect of these ambiguities on the performance of the NLP tools is expected to be minimal.

9. `http://wordlist.sourceforge.net/`

reduced rest; no runway signs and busy doing last minute commuter paper work changes; all contributed to the runway incursion. 12 hour day 6 hour flight time.

We ran the POS tagger, CRFTagger, on this processed text and did not observe any errors. For example, the tagged version of the aforementioned two sentences are:

1. TAXIING/VBG from/IN the/DT ramp/NN at/IN LAF/NNP at/IN night/NN ./.

2. made/VBN a/DT wrong/JJ turn/NN and/CC crossed/VBD runway/NN 10/28/CD ;/: the/DT active/JJ at/IN the/DT time/NN ./.

Both sentences have been correctly tagged. However, our case restoration method is arguably too simplistic. Hence, to determine if we need to perform more fine-grained case restoration, we sought a measure of how much would we gain from accurately restoring the case of the words in the sentences over the present heuristic method. To check this, we randomly picked 100 sentences from the corpus. We first ran the POS tagger on these sentences after they were case-restored by the aforementioned heuristic case restoration method. Then, we manually corrected the capitalization of these sentences and re-ran the POS tagger on the case-restored sentences. When the tags thus generated were compared, we found 99.7% agreement, which means that we are not likely to gain much in terms of POS tagging accuracy from correctly case restored text than the heuristically case restored text. Of the five differences out of 2049 words, three were NNPs mislabeled as NNs, which essentially has no effect on outcomes of our research. Therefore, the marginal utility from applying more sophisticated case restoration methods does not seem enough to justify the additional effort necessary, and we limit our preprocessing step to the expansion of abbreviations and acronyms followed by the heuristic case restoration procedure described above. The complete flow of preprocessing is shown in Figure 1.

## 2.3 Human Annotation Procedure

Recall that we need reports labeled with the shaping factors for training the cause identification classifiers and testing the performance of our two approaches to cause identification. Additionally, in order to learn a semantic lexicon via bootstrapping, we need a small set of seed words and phrases related to each shaping factor as a starting point. As a result, after performing language normalization, we performed *two* types of annotations: (1) labeling a set of reports with shaping factors, and (2) identifying a set of seed words and phrases from the reports. The annotation procedure is described in more detail in the following sections.

### 2.3.1 ANNOTATING REPORTS WITH SHAPING FACTORS

While NASA has previously developed a heuristic approach to tackle the cause identification task (Posse et al., 2005), this approach was evaluated on only 20 manually annotated reports, which is far from satisfactory as far as establishing a strong baseline method is concerned. Thus we decided to annotate a set of reports ourselves for evaluating our automatic cause identification methods.

Out of the complete set of 140,599 reports, we chose a random set of 1333 reports for annotation. This subset was divided into two parts. The first part, consisting of 233 reports, was annotated by two persons (one undergraduate student and one graduate student). For each report, they were asked to answer the following question:

> Which shaping factor(s) were responsible for the incident described in the report?

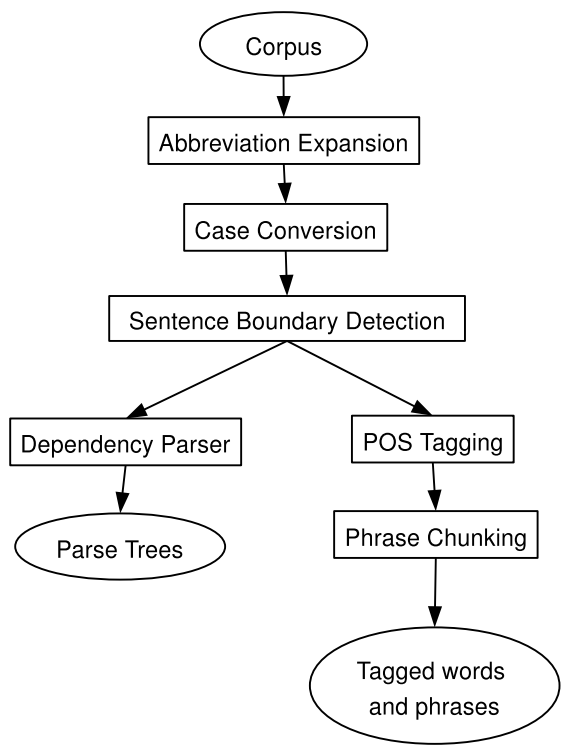

Figure 1: Flow chart of text preprocessing

Our annotators were trained in a similar way as those who labeled the 20 reports used in the evaluation by the NASA researchers (see Posse et al., 2005). Specifically, as background reading, the annotators were referred to the works of Posse et al. and Ferryman et al. (2006), both of which describe the shaping factors, and also give some examples of the words and phrases that indicate the influence of the shaping factors on the described incidents. The definitions of the shapers are repeated in Section 2.1. Following Posse et al.'s method, our annotators were explicitly instructed to adhere to these definitions as much as possible when annotating the reports with shaping factors. After the annotations were completed, the inter-annotator agreement was computed using the Krippendorff's (2004) $\alpha$ statistics as described by Artstein and Poesio (2008), using the Measuring Agreement on Set-valued Items (MASI) scoring metric (Passonneau, 2004). The observed inter-annotator agreement, $\alpha$, in this case was found to be 0.72, which indicates reliable agreement. Out of the 233 reports, they completely agreed on the annotations of 80 reports, completely disagreed on 100 reports and partially agreed on 53 reports. The annotators were then asked to discuss the discrepancies. During the discussion, it was found that the discrepancies could be primarily attributed to the vagueness of the descriptions of the shaping factors in Posse et al.'s paper, some of which were interpreted differently by the two annotators.

The annotators then agreed on how the descriptions of the shapers should be interpreted, and resolved all the differences in their annotation. After the discussion, the remaining 1100 reports were annotated by one of the annotators. The other annotator was also asked to annotate a subset of these reports (100 reports) for cross-verification purpose[10], and the inter-annotator agreement, $\alpha$, in this case was observed to be 0.66. The 1333 reports annotated by the first annotator were divided into three sets: a *training* set (233 reports) for training the cause identification classifiers, a held-out *development* set (100 reports) for parameter tuning, and a *test* set (1000 reports) for evaluating the performance of our approaches to cause identification. The distribution of the shaping factors in the training, development and test sets are shown in the second, third and fourth columns of Table 1.

### 2.3.2 EXTRACTING SEED WORDS AND PHRASES

In a separate process, the first author went through the first 233 reports that both annotators worked on, and selected words and phrases relevant to each of the shaping factors. His judgment of whether a word or phrase is relevant to a shaping factor was based on a careful reading of the description of the shaping factors in the works of Posse et al. (2005) and Ferryman et al. (2006), as well as the example seed words selected by the NASA experts that were shown in these two papers. The specific task in this case was:

> In each report, is there any word or phrase that is indicative of any of the shaping factors? If there is, then identify it and assign it to the appropriate shaping factor.

Note that these seed words and phrases were chosen without regard to the shaping factor annotation of the document; they were picked on the possibility of their being relevant to the respective shaping factors. The number of seed words and phrases for each shaping factor is shown in the last column of Table 1. As we can see, 177 seed words and phrases were manually selected from the 233 training reports. For completeness, we also show all the seed words and phrases extracted from these reports in Appendix A. To facilitate further research on this topic, the annotated data we have used in this research is made available at `http://www.utdallas.edu/~maa056000/asrs.html`.

Since there is no gold standard against which we can compare this list of annotated words and phrases, it is difficult to directly compute its precision. However, to get a rough idea of its precision, we asked one of the annotators to examine the list and identify all and only those words and phrases in the list that he believes are correct. There was disagreement over only one word. This yields a precision of 99.44%, which provides suggestive evidence that the annotation is fairly reliable. These manually identified words and phrases were used by our baseline cause identification system (see Section 3) and also served as seeds for our semantic lexicon learning procedure (see Section 4).

---

10. It is a fairly standard procedure in NLP research to cross-annotate only a subset of the data when complexity and cost of individual annotation is high. See the works of Zaidan, Eisner, and Piatko (2007) and Kersey, Di Eugenio, Jordan, and Katz (2009), for instance.

Table 1: Distribution of shaping factors in the training, test and development sets

| Shaping factor | Reports in training set | Reports in test set | Reports in development test set | Seed words |
|---|---|---|---|---|
| Attitude | 17 | 30 | 5 | 8 |
| Communication Environment | 11 | 90 | 18 | 5 |
| Duty Cycle | 9 | 26 | 3 | 10 |
| Familiarity | 12 | 50 | 8 | 9 |
| Illusion | 1 | 2 | 0 | 1 |
| Other | 36 | 217 | 36 | 8 |
| Physical Environment | 43 | 265 | 40 | 45 |
| Physical Factors | 10 | 35 | 3 | 8 |
| Preoccupation | 25 | 110 | 10 | 9 |
| Pressure | 5 | 30 | 3 | 10 |
| Proficiency | 43 | 247 | 23 | 12 |
| Resource Deficiency | 112 | 507 | 33 | 47 |
| Taskload | 6 | 29 | 7 | 2 |
| Unexpected | 3 | 10 | 1 | 3 |
| Total | 233 | 1000 | 100 | 177 |

## 3. Baseline System For Cause Identification

As discussed in the introduction, the goal of our research is to label the incident reports with the shaping factors that caused the incidents. To evaluate the performance of our cause identification methods, we need a baseline that uses the same amount of training data as all the methods described in this paper and performs reasonably well on the test set. Given that cause identification is a relatively new and under-investigated task, no standard baseline has been adopted for this task. In fact, to our knowledge, the only related works on cause identification for the aviation safety domain were conducted by the researchers at NASA (see Posse et al., 2005; Ferryman et al., 2006). As a result, we construct a baseline system motivated by Posse et al.'s work. Specifically, the baseline takes as input a set of seed words and phrases manually collected for each of the shaping factors (see Section 2.3.2), and labels a report with the *Occurrence Heuristic*: for each seed word and phrase found in the report, the baseline annotates the report with the shaping factor associated with the seed. For example, "11 hour duty day" is a seed phrase associated with the shaping factor *Duty Cycle*. Then, the Occurrence Heuristic will label any report that contains the phrase "11 hour duty day" with *Duty Cycle*. This approach is simple but attractive because (1) it does not need any training, (2) it can be evaluated very easily, by searching for the seed words in the narrative of the report being labeled, and (3) a report can potentially be labeled with more than one shaping factors. If the seed words and phrases are indeed relevant to their respective shaping factors, then they should identify the reports related to the shaping factors with a high degree of precision.

## 4. Semantic Lexicon Learning

As described in Section 3, the baseline uses the seed words and phrases manually extracted from 233 reports in combination with the Occurrence Heuristic to label the reports with shaping factors. However, the reports used for evaluation may not contain exactly the same words and phrases, but they may contain different variations, synonyms, or words and phrases that are semantically similar to the seed words and phrases. Thus the baseline may not be able to label these reports correctly by only looking for the words and phrases in the seed words list.

To address this potential problem, we propose to use semantic lexicon learning algorithms to learn more words and phrases semantically similar to the seed words and phrases from the reports corpus containing narratives from 140,599 reports. Using a weakly supervised bootstrapping algorithm may allow us to learn a large number of useful words and phrases from the corpus that would have required huge amounts of human effort had it been done manually. Below, we first describe the general bootstrapping approach in Section 4.1. Then, in Section 4.2, we describe the Basilisk framework for learning the semantic lexicon from an unannotated corpus (Thelen & Riloff, 2002). Finally, in Section 4.3, we discuss our modifications to the Basilisk framework.

### 4.1 Weakly Supervised Lexicon Learning

As mentioned earlier, we employ a weakly supervised bootstrapping approach for building the semantic lexicon. We use the manually extracted seed words and phrases for each shaping factor (described in Section 2.3.2) to create the initial semantic lexicon. Then we select words and phrases from the unannotated reports that are semantically similar to the words already appearing in the semantic lexicon. The reports in the corpus do not need to be labeled with shaping factors. The semantic similarity between two words is measured using features extracted from the corpus for each word. This process is repeated iteratively: in each iteration, a certain number of words are added to the semantic lexicon, and the words in this augmented lexicon are used as the seeds for the following iteration. This process is shown in Figure 2.

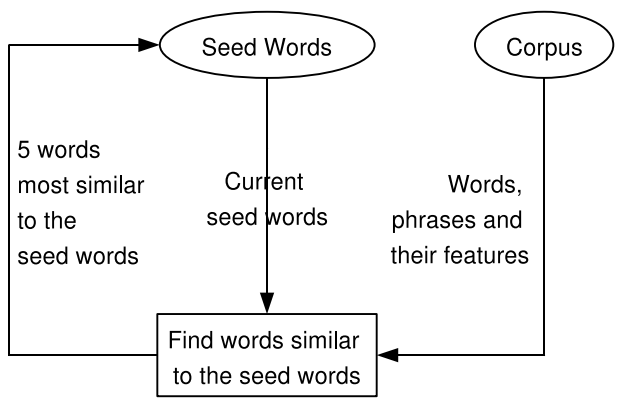

Figure 2: Flow chart of the lexicon learning procedure

## 4.2 Basilisk Framework

Basilisk (Bootstrapping Approach to SemantIc Lexicon Induction using Semantic Knowledge) is an instantiation of the aforementioned generic semantic lexicon learning framework (Thelen & Riloff, 2002). The Basilisk framework works by first identifying all the *patterns* for extracting all the noun phrases in the corpus that appear in one of three syntactic roles: subject, direct object, or prepositional phrase object. For example, as discussed by Thelen and Riloff, in the sentence *"**John was arrested** because he <u>collaborated with **Smith**</u> and <u>murdered **Brown**</u>"*, the extraction patterns are *"<subject> was arrested"*, which extracts *John*, *"murdered <object>"* which extracts *Brown* and *"collaborated with <pp_object>"* which extracts *Smith*. Then, for each semantic category $S_k$, a *pattern pool* is constructed with patterns that tend to extract words in $S_k$. To measure the tendency of a pattern $P_j$ to extract words in $S_k$, the $R \log F$ metric is used, which is defined as:

$$R \log F (P_j) = \frac{F_j}{N_j} \times \log (F_j) \tag{1}$$

Here, $F_j$ is the number of (distinct) words in $S_k$ that pattern $P_j$ extracts, and $N_j$ is the total number of (distinct) words in the corpus that $P_j$ extracts. This metric is high for both high precision patterns (i.e., patterns that extract primarily words in $S_k$) and high recall patterns (i.e., patterns that extract a large number of words in $S_k$). At each iteration $i$, the top $(20 + i)$ patterns (in terms of their $R \log F$ scores) are put into the pattern pool for $S_k$. Depleted patterns (i.e., patterns that have all their extracted words already in the semantic lexicon) are not considered in this step. Then, the head nouns of all the phrases extracted by the resulting patterns in the pattern pool are put into the *word pool* of $S_k$.

Next, a subset of the words in the word pool is selected to be added to the seed words list. Those words from the word pool are chosen that are most relevant to $S_k$. More specifically, for each word $W_i$ in the word pool for $S_k$, first the *AvgLog* score is calculated, which is defined as follows:

$$AvgLog (W_i, S_k) = \frac{\sum_{j=1}^{WP_i} \log_2 (F_j + 1)}{WP_i} \tag{2}$$

Here, $WP_i$ is the number of patterns that extract word $W_i$, and for each pattern $P_j$ that extracts $W_i$, $F_j$ is the number of words extracted by $P_j$ that belong to $S_k$. Then, for each semantic category $S_k$, five words are chosen that have the highest *AvgLog* score for the category $S_k$.

For multi-category learning, Thelen and Riloff (2002) experimented with different scoring metrics and reported that they achieved the best performance by calculating the *diff* score for each word. For a given word in the word pool for a semantic category, the *diff* score takes into consideration what score this word gets for the other categories, and returns a score based on the word's score for this semantic category relative to the other categories. More precisely, the *diff* score is defined as follows:

$$diff (W_i, S_k) = AvgLog (W_i, S_k) - \max_{l \neq k} (AvgLog (W_i, S_l)) \tag{3}$$

Here, $S_k$ is the semantic category for which $W_i$ is being evaluated. Thus the *diff* score is high if there is strong evidence that $W_i$ belongs to semantic category $S_k$ but little evidence that it belongs to the other semantic categories. For each semantic category, the *diff* score is calculated for each word in the category's word pool, and the top five words with the highest *diff* score are added to the lexicon for that category. Two additional checks are made at this stage: (1) if a word in the word pool has been added to some other category in an earlier iteration, that word is discarded, and (2) if the same word is found in more than one word pool then it is added to the category for which it has the highest score[11]. When this is completed for all the semantic categories, the iteration ends, and the next iteration begins with the augmented lexicon.

## 4.3 Modifications to the Basilisk Framework

As we will see later in this subsection, an analysis of the framework reveals that in some cases the words selected by Basilisk may not be the most relevant ones. For this reason, we propose three algorithmic modifications to the Basilisk framework: (1) using a new semantic similarity measure, (2) merging the word pools to one single pool for assigning words to the semantic categories, and (3) imposing minimum support and maximum generality criteria on patterns and words added to the pattern pools and the word pools. In addition, we propose one linguistic modification, in which we employ a type of feature that can be computed in a robust manner from the words and phrases in the corpus, namely, the N-gram features. The rest of this subsection discusses these modifications.

### 4.3.1 MODIFICATION 1: NEW SEMANTIC SIMILARITY MEASURE

As seen in Section 4.2, the Basilisk framework uses the *AvgLog* scoring function to measure the semantic similarity between words. The *diff* score for multi-category learning also uses the *AvgLog* function to compute the evidence for a word belonging to a semantic category relative to the other categories. However, a closer examination of the *AvgLog* function shows that it may not be able to properly predict semantic similarity under all circumstances. To understand the reason, let us first make the following observations: if pattern $P_j$ occurs 1000 times, but extracts words in category $S_k$ only 5 times, it is unlikely that $P_j$ is strongly related to $S_k$. Similarly, if word $W_i$ occurs 1000 times, but is extracted by pattern $P_j$ only 5 times, $P_j$ should have small influence on the classification of $W_i$. However, the *AvgLog* score will not be able to take these factors into consideration, precisely because it considers only the absolute number of semantic category members extracted by the patterns that extract the word but not the frequency of extraction. To see why this is the case, let us consider the word $W_i$ that is extracted by three patterns $P_1$, $P_2$ and $P_3$, with the frequencies as shown in Table 2. If each of $P_1$, $P_2$ and $P_3$ extract five distinct seed words, then the *AvgLog* score for the word $W$ would be 2.32, irrespective of the fact that the patterns actually extract a word in the seed words list only a tiny fraction of their occurrence in the corpus. $P_1$ extracts a seed word 5% of its occurrence, $P_2$ does so 1% time, and $P_3$, the pattern that extracts $W$ most often, extracts a lexicon word only 0.5% of the times it appears in the text. Clearly, the patterns would not suggest that $W_i$ is related to the semantic category, yet it gets a good score.

---

11. This approach effectively assumes that each word can belong to at most one category. This is a reasonable assumption in this specific task since the shaping factors have very distinct meanings.

Table 2: Illustration of the problem with *AvgLog*: How unrelated words may have a high similarity score. Here $W_i$ is a word that appears in the corpus and is extracted by the patterns $P_1$, $P_2$ and $P_3$

| Patterns that extract $W_i$ | $P_1$ | $P_2$ | $P_3$ |
|---|---|---|---|
| Number of times $W_i$ is extracted by the pattern $P_j$ | 10 | 20 | 70 |
| Number of times pattern $P_j$ occurs in the text | 100 | 500 | 1000 |
| Number of times a word in category $S_k$ is extracted by the pattern $P_j$ | 5 | 5 | 5 |
| Number of category words extracted by the pattern $P_j$ | 5 | 5 | 5 |
| $\log_2(F_j + 1)$ | 2.32 | 2.32 | 2.32 |
| $AvgLog(W_i)$ | 2.32 | | |

Keeping this in mind, we propose our probabilistic metric, *SemProb*, which computes the probability that the word $W_i$ belongs to the semantic category $S_k$ given that it is extracted by the patterns $P_1$, $P_2$, ..., $P_n$. More specifically, *SemProb* is calculated as follows:

$$\begin{aligned} SemProb(W_i, S_k) &= Prob(S_k|W_i) \\ &= \sum_{P_j} Prob(S_k|P_j) \times Prob(P_j|W_i) \end{aligned} \quad (4)$$

In other words, *SemProb* assumes that the semantic category $S_k$ and the word $W_i$ are conditionally independent given $P_j$, a pattern that extracts $W_i$. The probabilities in this equation are estimated using maximum likelihood estimation from the corpus. Specifically, to compute $Prob(P_j|W_i)$, we divide the number of times $P_j$ extracts $W_i$ in the corpus by the total number of times that $W_i$ appears in the corpus. To compute $Prob(S_k|P_j)$, we divide the number of times $P_j$ extracts a word in the semantic category $S_k$ by the total number of times $P_j$ appears in the corpus. For a given word $W_i$ and a given semantic category $S_k$, the sum of the products of these two quantities over all the patterns that extract $W_i$ gives the probability of category $S_k$ given word $W_i$. This method does not suffer from the problem faced by *AvgLog* since it depends on the probability of the word being extracted by the patterns and the patterns' probability of extracting words in the category. For the same example in Table 2, the *SemProb* metric for the word $W_i$ is 0.0105, illustrating how low the probability of $W_i$'s belonging to the semantic category $S_k$ is. The details are given in Table 3.

### 4.3.2 MODIFICATION 2: COMMON WORD POOL

Since we have to compute Eqn (4) for every word in the word pool for each of the categories and assign the word to the semantic category for which the probability is highest, we change the framework so that we have only one common *word pool* for all the semantic categories.

Table 3: Illustration of the effectiveness of *SemProb*: How unrelated words get low similarity score.

| Patterns that extract $W_i$ | $P_1$ | $P_2$ | $P_3$ |
|---|---|---|---|
| Number of times that $W_i$ is extracted by the pattern $P_j$ | 10 | 20 | 70 |
| Number of times pattern $P_j$ occurs in the text | 100 | 500 | 1000 |
| Number of times a word in category $S_k$ is extracted by the pattern $P_j$ | 5 | 5 | 5 |
| $Prob\,(W_i is\ extracted\ by\ P_j)$ | 0.1 | 0.2 | 0.7 |
| $Prob\,(P_j\ extracts\ a\ word\ in\ S_k)$ | 0.05 | 0.01 | 0.005 |
| $Prob\,(W_i is\ extracted\ by\ P_j) \times Prob\,(P_j\ extracts\ a\ word\ in\ S_k)$ | 0.005 | 0.002 | 0.0035 |
| $SemProb\,(W_i, S_k) = Prob\,(W_i\ belongs\ to\ semantic\ category\ S_k)$ | 0.0105 | | |

We still have separate pattern pools for different semantic categories, but the words related to patterns in the pattern pools will be put into the same common word pool, and allocated to the most probable semantic category from there. If there are separate word pools for each semantic category, then we have to add a fixed number of words to each category in each iterations. Such a constraint may undesirably cause a word to be added to a category that is not the most likely. However, since we have only one word pool after our modification, we do not have the constraint that we have to add a fixed number of words to each category, and we can assign each word to its most likely category. Thus the number of words added to different categories may vary in the same iteration.

### 4.3.3 MODIFICATION 3: MINIMUM SUPPORT AND MAXIMUM GENERALITY

There are some scenarios in which the *SemProb* metric can produce undesirable results. For example, consider a very infrequent word $W_i$ that occurs in the entire corpus exactly once. Assume that pattern $P_j$, which extracts $W_i$, extracts words in semantic category $S_k$ with 70% probability. So, according to *SemProb*, the probability that $W_i$ belongs to $S_k$ becomes 70%. However, this is not sufficient evidence for $W_i$ to belongs $S_k$. Such cases not being too uncommon, we have imposed a *minimum word frequency* constraint on the words that are put into the word pool, so that words that appear less than a certain number of times are not considered. A pattern that appears too infrequently in the corpus can also lead to such a problem. Consider a very infrequent pattern, $P_j$, that appears exactly twice in the corpus and extracts two words. If one of these words happen to be a seed word, then the other word will have a 50% probability to belong to the category of the seed word and $P_j$ will have $R \log F$ value of 0.5. However, since $P_j$ is so infrequent, it does not convey a good evidence for membership in the semantic category, and we should not allow $P_j$ to put words into the word pool. Therefore, we disallow such low frequency patterns from being included in the pattern pool by adding the constraint that the patterns put into the pattern pool must also have a *minimum pattern frequency*. Besides these two constraints imposed on the frequency of occurrence of the words and the patterns, we employ two additional constraints. The first is the *maximum pattern generality* constraint: motivated by Rychlý and Kilgarriff (2007), we remove from consideration patterns that are too general (i.e., patterns that extract too many words), by imposing an upper limit on the number of distinct words that a pattern to be added to a pattern pool can extract. The second is the *maximum word frequency* constraint: since content-bearing words are likely to have a lower frequency (see Davidov & Rappoport, 2006), we impose an upper limit on the maximum number of times a word appears in the corpus. The four thresholds associated with these four frequency-based constraints will be tuned automatically using the held-out development set.

### 4.3.4 Modification 4: N-gram Patterns

In addition to the parse-tree-based subject-verb and verb-object patterns already employed by Basilisk, we also employ N-gram-based extraction patterns, with the goal of more robustly capturing the context in which the words appear. We construct N-gram extraction patterns as follows. For each noun and adjective, $X$, in the corpus, we create two N-gram patterns for extracting $X$: (a) the preceding N words + $\langle X \rangle$, and (b) $\langle X \rangle$ + the succeeding N words. For example, in the sentence "... a solid line of thunderstorms was detected ...", the bigram patterns for "thunderstorms" would be: "line of $\langle X \rangle$" and "$\langle X \rangle$ was detected". The complete sentence is "approaching the *ATL area* a *solid line* of *thunderstorms* was detected in the *vicinity* of the *airport*," and the words and their extracting bigram patterns would be:

- *ATL*: approaching the $\langle X \rangle$, $\langle X \rangle$ area a

- *area*: the ATL $\langle X \rangle$, $\langle X \rangle$ a solid

- *solid*: area a $\langle X \rangle$, $\langle X \rangle$ line of

- *line*: a solid $\langle X \rangle$, $\langle X \rangle$ of thunderstorms

- *thunderstorms*: line of $\langle X \rangle$, $\langle X \rangle$ was detected

- *vicinity*: in the $\langle X \rangle$, $\langle X \rangle$ of the

- *airport*: of the $\langle X \rangle$

In addition to constructing N-gram patterns for extracting words, we also construct N-gram patterns for extracting *phrases*. To do so, we first remove articles (a, an, the) and possessive pronouns and adjectives (e.g., my, his) from the beginning of the phrases in the corpus. For each noun phrase and adjective phrase, $X$, that appears in the corpus, we create two N-gram patterns for extracting $X$: (a) The preceding N words + $\langle X \rangle$, and (b) $\langle X \rangle$ + the succeeding N words. For example, from the sentence "this was the last of *5 legs* and approaching *the end* of *an 8 hour duty day* and *7 hour hard time flying day*", we would extract the following phrases with the following bigram patterns:

- *5 legs*: last of $\langle X \rangle$, $\langle X \rangle$ and approaching

- *end*: and approaching $\langle X \rangle$, $\langle X \rangle$ of an

- *8 hour duty day*: end of $\langle X \rangle$, $\langle X \rangle$ and 7

- *7 hour hard time flying day*: day and $\langle X \rangle$

Thus we use three types of patterns in our experiments: bigram patterns for extracting words, bigram patterns for extracting phrases, and parse-tree-based subject-verb and verb-object patterns. All these patterns were generated from the reports corpus generated by combining the narratives of the 140,599 unlabeled reports described in Section 2.2. As we will see, not all three types of patterns are beneficial to use as far as performance is concerned. In Section 6, we will show how to automatically select the best subset of patterns to use based on the development set.

## 5. Semantic Lexicon-Based Approaches to Cause Identification From ASRS Reports

We investigate a heuristic-based approach and a learning-based approach to cause identification, both of which exploit information provided by an automatically acquired semantic lexicon. This section describes the details of these two approaches.

### 5.1 Heuristic-Based Approach

The heuristic-based approach operates in essentially the same way as the baseline cause identification system described in Section 3, where the Occurrence Heuristic is used to label a report with shaping factors. The only difference is that the words and phrases used by the Occurrence Heuristic in the baseline are manually identified, whereas those in our heuristic-based approach are acquired by our Modified Basilisk procedure.

### 5.2 Learning-Based Approach

Our learning-based approach to the cause identification problem is to recast it as a classification task. Note that we have a multi-class multi-labeled classification task: there are 14 classes and each report can be labeled with more than one class. A number of approaches have been proposed to tackle multi-class multi-labeled classification tasks. In the rest of this section, we describe the three existing approaches to multi-class multi-labeled text classification that we explore in our experiments (Section 5.2.1), and provide an overview of the theory of Support Vector Machines (SVMs), the underlying learning algorithm we use to train classifiers employed by these three approaches (Section 5.2.2).

#### 5.2.1 THREE APPROACHES TO MULTI-CLASS MULTI-LABELED TEXT CLASSIFICATION

**One-Versus-All.** In this approach, we train one binary classifier for each shaping factor $S_k$ to determine whether a report will be labeled with $S_k$. More specifically, we follow the One-Versus-All classification scheme: for a given $S_k$, the reports in the training set that contains $S_k$ in its set of labels (assigned by the annotator) are the positive instances for the binary classifier and the rest of the reports in the training set are the negative instances. After training, we apply the classifiers to a report in the test set independently of other reports, and label the report with each $S_k$ for which the corresponding classifier classifies the report as positive. Thus we convert cause identification to a multi-class multi-labeled document classification task.

While any learning algorithm can be used in principle to train classifiers for this One-Versus-All scheme, we use Support Vector Machines[12] for training and testing the classifiers, primarily due to its successes in various text classification tasks. Each classifier is trained with two types of features: (1) unigrams and bigrams from the report narratives, and (2) words and phrases from the semantic lexicon. The feature values are TF*IDF values.

While our shaping factor-labeled data set of 1333 reports is substantially larger than the set of 20 reports annotated by the NASA researchers (see Section 1), it is arguably fairly small from a machine learning perspective. Hence, it is conceivable that the performance of our SVM classifiers would be limited by the small size of the training data. As a result, we investigate whether we can improve the One-Versus-All approach using a *transductive* SVM, which is a version of the inductive SVM described above that attempts to improve classifier performance by combining both labeled and unlabeled data (see Section 5.2.2 for an overview of transductive learning). For our cause identification task, the unlabeled reports in the test set serve as unlabeled data in the transductive learning procedure.

**MetaLabeler.**   As our second approach, we employ *MetaLabeler* (Tang, Rajan, & Narayanan, 2009) for classifying multi-class multi-labeled text data. Here, a model is first learned that predicts the number of labels that an instance may have. In addition, a set of binary classifier models, one for each possible label, are learned to predict the likelihood of each label for an instance. When an instance is classified, the first model predicts $K$, the number of possible labels for that instance, and from the output of the second set of classifiers, $K$ labels are chosen with the highest likelihood for that instance.

In our implementation of this approach, the first model is learned using $\text{SVM}^{multiclass}$, which is an implementation of multi-class SVM described by Crammer and Singer (2002)[13]. The second set of classifiers are the same set described in Section 5.2.2. But in this case, for a given instance $\mathbf{x}$, the decision functions $f(x) = \mathbf{w} \cdot \mathbf{x} - b$ for each of the classifiers are evaluated, and the positive decision values are sorted. Then the top $K$ labels corresponding to the highest values of the decision functions are assigned to the instance. Both the multiclass classifier and the set of binary classifiers are trained using the same types of features as in the One-Versus-All approach, namely unigrams and bigrams from the reports, and words and phrases from the semantic lexicon. The feature values are also the same as in One-Versus-All approach, namely TF*IDF values.

**Ensembles of Pruned Sets.**   In the *Pruned Sets* approach (Read, Pfahringer, & Holmes, 2008), the multi-class multi-label text classification problem is transformed into a multi-class *single*-label text classification problem by selecting a subset of the label combinations most frequently occurring in the dataset and assigning a unique pseudo-label to each chosen label combination.

The first step in this algorithm is to choose the label sets for training. In this step, those label sets are chosen that meet the minimum frequency requirement in the training set. Using the minimum frequency constraint prunes away infrequently occurring label sets that have frequency less than $p$, leaving only label combinations that are frequent and thus more important. The training instances that are labeled with the pruned label sets are also removed from the training set. The minimum cardinality parameter, $b$, is then used to reintroduce some of the pruned instances back to the training set in order to minimize the information loss from the pruning process. First the label sets of the rejected instances are broken down into smaller subsets of at least size $b$. Then those new subsets that have frequency higher than $p$ are reintroduced, and the pruned training instances whose label sets are supersets of these newly accepted label sets are reinstated into the training set. The role of the parameter $b$ in this case is to ensure that not too many such instances with small label sets are put back, because that will cause the average number of labels to reduce, resulting in smaller number of labels per instance at classification time.

---

12. As implemented in the $\text{SVM}^{light}$ software package by Joachims (1999)
13. Available at `http://svmlight.joachims.org/svm_multiclass.html`

The next step is to learn classifiers on the selected label sets. First, each accepted label set is assigned a unique pseudo-label, thus transforming the multi-label classification problem into a single-label classification problem. Then an ensemble of $M$ classifiers is learned to predict these pseudo-labels given an instance (using the same multi-class SVM implementation as in MetaLabeler), where each classifier in the ensemble is trained on a different random sample of the training data. Since (1) the label sets for training the classifiers represent only a subset of all the label combinations present in the original training data and (2) the test data may contain label combinations that are not present in the training data, having an ensemble of classifiers allows the system to generate label combinations not observed at training time. For example, let the label combinations $\{l_1, l_3\}$ and $\{l_2, l_3\}$ be present in the training data. Then, if one classifier in the ensemble labels a test instance with $\{l_1, l_3\}$ and another classifier in the ensemble labels the same instance with $\{l_2, l_3\}$, then that instance may be labeled with $\{l_1, l_2, l_3\}$ (depending on the actual voting policy in effect at classification time) even if this combination is not present in the training data. The classifiers in the ensemble are built using the same two types of features as the One-Versus-All approach, namely unigrams and bigrams from the reports and words and phrases from the semantic lexicon learned by our modified Basilisk framework.

Finally, when classifying an instance, each of the $M$ classifiers assigns one pseudo-label to the instance. These pseudo-labels are then mapped back to the original label combination and the vote for each actual label is counted and normalized by dividing by the number of classifiers, $M$, in order to bring the prediction for each possible label to the range between 0.0 and 1.0. Then a threshold $t$ is used such that each label that has a prediction value greater than or equal to $t$ is assigned to the instance. This scheme is used to make it possible to assign label combinations unseen at training time to the test instances.

### 5.2.2 An Overview of Support Vector Machines

SVMs have been shown to be very effective in text classification (Joachims, 1999). Below we describe two versions of SVMs: (1) inductive SVMs, which learn a classifier solely from labeled data, and (2) transductive SVMs, which learn a classifier from both labeled and unlabeled data.

**Inductive SVMs.**   Given a training set consisting of data points belonging to two classes, an inductive SVM aims to find a separating hyperplane that maximizes the distance from the separating hyperplane to the nearest data points. These nearest data points act as the *support vectors* for the plane.

More formally, let $D$ be the data set with $m$ data points where

$$D = \{(\mathbf{x}_i, c_i) \,|\, \mathbf{x}_i \in \mathbf{R}^n, c_i \in \{-1, 1\}, 1 \leq i \leq m\} \tag{5}$$

Each point $\mathbf{x}_i$ is represented as an $n$-dimensional vector and is associated with a class label $c_i$. The inductive SVM classifier attempts to find a hyperplane $\mathbf{w} \cdot \mathbf{x} - b = 0$ that is at the maximum distance from the nearest data points of opposite labels. This hyperplane would be in the middle of the two hyperplanes containing the support vectors of each class. These two hyperplanes are $\mathbf{w} \cdot \mathbf{x} - b = 1$ and $\mathbf{w} \cdot \mathbf{x} - b = -1$, and their distance is $\frac{2}{|\mathbf{w}|}$. Therefore, the desired separating hyperplane can be found by solving the following quadratic programming optimization problem:

$$\begin{aligned} \text{Minimize} \quad & \frac{1}{2}|\mathbf{w}|^2 \\ \text{subject to} \quad & c_i\left(\mathbf{w} \cdot \mathbf{x}_i - b\right) \geq 1, 1 \leq i \leq m \end{aligned} \tag{6}$$

However, in practice many classes are not linearly separable. To handle these cases, a set of *slack variables* is used to represent the misclassification of point $\mathbf{x}_i$. Then the problem becomes:

$$\begin{aligned} \text{Minimize} \quad & \frac{1}{2}|\mathbf{w}|^2 + C \sum_i \xi_i \\ \text{subject to} \quad & c_i\left(\mathbf{w} \cdot \mathbf{x}_i - b\right) \geq 1 - \xi_i, \, \xi_i > 0, 1 \leq i \leq m \end{aligned} \tag{7}$$

where the $\xi_i$ are additional variables representing training errors and $C$ is a constant representing trade-off between training error and margin. More details can be found in Cortes and Vapnik (1995). In our experiments, we use the *radial basis function (RBF)* kernel, where every dot product is replaced by the function $k\left(\mathbf{x}, \mathbf{x}'\right) = \exp\left(-\gamma |\mathbf{x}, \mathbf{x}'|^2\right)$, for $\gamma > 0$. In addition, both $\gamma$ and $C$ are chosen by cross-validation on the training set.

**Transductive SVMs.** In the transductive setting, in addition to the set of labeled data points, we also exploit a set of unlabeled data points, $T = \{\mathbf{x}_i^* | \mathbf{x}_i^* \in \mathbf{R}^n, 1 \leq i \leq k\}$, that are taken from the test set. As described by Joachims (1999), the goal is then to minimize the expected number of classification errors over the test set. The expected error rate is defined in Vapnik (1998) as follows:

$$R(L) = \int \frac{1}{k} \sum_i \Theta\left(h_L\left(\mathbf{x}_i^*\right), c_i^*\right) dP(\mathbf{x}_1, c_1) \ldots dP(\mathbf{x}_k^*, c_k^*) \tag{8}$$

where $L = D \cup T$, $h_L$ is the hypothesis learned from $L$, and $\Theta(a, b)$ is zero if $a = b$ and one otherwise. The labeling $c_i^*$ of the test data and the hyperplane that maximizes the separations of both training and testing positive and negative instances are found by solving the following quadratic programming optimization problem, which is a modified version of Eqn (7):

$$\begin{aligned} \text{Minimize} \quad & \frac{1}{2}|\mathbf{w}|^2 + C \sum_i \xi_i + C^* \sum_j \xi_j^* \\ \text{subject to} \quad & c_i\left(\mathbf{w} \cdot \mathbf{x}_i - b\right) \geq 1 - \xi_i, \, \xi_i > 0, \, 1 \leq i \leq m \\ & c_j^*\left(\mathbf{w} \cdot \mathbf{x}_j^* - b\right) \geq 1 - \xi_j^*, \, \xi_j^* > 0, \, 1 \leq j \leq k \end{aligned} \tag{9}$$

Similar to the inductive SVM in Section 5.2.2, we use the RBF kernel in our experiments involving the transductive SVM.

## 6. Evaluation

The goal of our evaluation is to study the effectiveness of our two approaches to cause identification, namely the semantic lexicon learning approach and the classification approach. We do so by testing the performance of the approaches on a randomly chosen set of reports that have been manually annotated with the shaping factors that caused the incidents described in them (Section 2.3.1). We start by describing the experimental setup (Section 6.1), followed by the baseline results (Section 6.2) and the performance of our two approaches (Sections 6.3 and 6.4). We then describe the experiment where we increase the amount of training data available to the classification approach and investigate how this impacts performance (Section 6.5). After that, we present an analysis of the errors of the best-performing approach (Section 6.6) and conduct additional experiments in an attempt to gain a better insight into the cause identification task that can help direct future research (Section 6.7). Finally, we present a summary of the major conclusions that we draw from the experiments (Section 6.8).

### 6.1 Experimental Setup

As described in Section 2.3, out of the 140,599 reports in the entire corpus, we have manually annotated 1333 incident reports with the shaping factors. We have used the first 233 of them to (1) manually extract the initial seed words and phrases for the semantic lexicon learning procedure, and (2) train classifiers for identifying shaping factors associated with a report. Of the remaining reports, we have used 1000 reports as test data and 100 reports as development data (for parameter tuning).

#### 6.1.1 EVALUATION METRICS

As mentioned in Section 2.1, there are 14 shaping factors, and a report may be labeled with one or more of these shaping factors. We evaluate the performance of our cause identification approaches based on how well the automatic annotations match the human annotations of the reports in the test set. For evaluation, we use *precision, recall* and *F-measure*, which are computed as described by Sebastiani (2002). Specifically, for each shaping factor $S_i, i = 1, 2, \ldots 14$, let $n_i$ be the number of reports in the test set that the human annotator has labeled with $S_i$, i.e., the number of true $S_i$-labeled reports in the test set. Further, let $p_i$ be the number of reports that an automatic labeling scheme $C_i$ has labeled with $S_i$, and let $tp_i$ be the number of reports that $C_i$ has labeled correctly with $S_i$. Then, for the shaping factor $S_i$, we have the following performance metrics:

- **Precision$_i$** is the fraction of reports that are really caused by shaping factor $S_i$ among all the reports that are labeled with $S_i$ by the labeling scheme.

$$Precision_i = \frac{tp_i}{p_i}$$

- **Recall$_i$** is the percentage of reports really caused by shaping factor $S_i$ that are labeled by the labeling scheme with the shaping factor $S_i$.

$$Recall_i = \frac{tp_i}{n_i}$$

Thus we obtain a measure of the labeling scheme's performance for each of the shaping factors. To obtain the overall performance of the labeling scheme, we sum these counts (i.e., $n_i$, $p_i$ and $tp_i$) over all shaping factors and compute the micro-averaged precision, recall and F-measure from the aggregated counts as described by Sebastiani and repeated as follows:

$$
\begin{aligned}
Precision &= \frac{\sum_i tp_i}{\sum_i p_i} \\
Recall &= \frac{\sum_i tp_i}{\sum_i n_i} \\
F\text{-}measure &= \frac{2 \times Precision \times Recall}{Precision + Recall}
\end{aligned}
$$

Thus for each labeling scheme we have one set of overall scores reflecting its performance over all classes.

### 6.1.2 Statistical Significance Tests

To determine whether a labeling scheme is better than another, we apply two statistical significance tests — McNemar's test (Everitt, 1977; Dietterich, 1998) and the stratified approximate randomization test (Noreen, 1989) — to test whether the difference in their performances is really statistically significant. McNemar's test compares two labeling schemes on the basis of errors (i.e., whether both the labeling schemes are making the same mistakes), and the stratified approximate randomization test compares the labeling schemes on F-measure. Both tests have been extensively used in machine learning and NLP literature. In particular, stratified approximate randomization is the standard significance test employed by the organizers of the Message Understanding Conferences to determine if the difference in F-measure scores achieved by two information extraction systems is significant (see Chinchor, 1992; Chinchor, Hirschman, & Lewis, 1993). Since we are ultimately concerned about the difference in F-measure scores between two labeling schemes in cause identification, our discussion of statistical significance in the rest of this section will be focused solely on the stratified approximate randomization test. For both tests, we determine significance at the level of $p < 0.05$.

## 6.2 Baseline System

Recall that we use as our baseline the heuristic method described in Section 3, where the Occurrence Heuristic is used to label a report using the seed words and phrases manually extracted from the 233 training reports. Results, shown in the Experiment 1 section of Table 4, are reported in terms of precision (P), recall (R), and F-measure (F). The last two columns show whether a particular automatic labeling scheme is significantly better than the baseline with respect to McNemar's test (MN) and stratified approximate randomization test (AR) [Statistical significance and insignificance are denoted by a ✓ and an ✗, respectively]. When evaluated on the 1000 reports in the test set, the baseline achieves a precision of 56.48%, a recall of 40.47% and an F-measure of 47.15%.

Table 4: Report labeling performance of different methods.

| Approach | Feature Set | P | R | F | MN | AR |
|---|---|---|---|---|---|---|
| **Experiment 1: Baseline** | | | | | | |
| Heuristic | Seed words | 56.48 | 40.47 | 47.15 | N/A | N/A |
| **Experiment 2: Semantic lexicon approach** | | | | | | |
| Heuristic | Lexicon from modified Basilisk | 53.15 | 47.57 | 50.21 | ✓ | ✓ |
| | Lexicon from original Basilisk | 49.23 | 42.78 | 45.78 | ✓ | ✗ |
| **Experiment 3: Supervised One-Versus-All classification approach** | | | | | | |
| SVM | Unigrams | 37.54 | 64.50 | 47.46 | ✓ | ✗ |
| | Unigrams and bigrams | 42.19 | 47.39 | 44.64 | ✓ | ✓ |
| | Lexicon words | 48.72 | 37.08 | 42.11 | ✓ | ✓ |
| | Unigrams and lexicon words | 37.05 | 65.96 | 47.45 | ✓ | ✗ |
| | Unigrams, bigrams, lexicon words | 51.19 | 36.59 | 42.68 | ✓ | ✓ |
| **Experiment 4: Transductive One-Versus-All classification approach** | | | | | | |
| SVM | Unigrams | 11.84 | 67.78 | 20.16 | ✓ | ✓ |
| | Unigrams and bigrams | 50.00 | 33.86 | 40.38 | ✓ | ✓ |
| | Lexicon from modified Basilisk | 42.83 | 30.64 | 35.73 | ✓ | ✓ |
| | Unigrams and lexicon words | 51.30 | 38.29 | 43.85 | ✓ | ✓ |
| | Unigrams, bigrams, lexicon words | 55.90 | 32.77 | 41.32 | ✗ | ✓ |
| **Experiment 5: MetaLabeler approach** | | | | | | |
| SVM | Unigrams | 58.80 | 16.63 | 25.92 | ✓ | ✓ |
| | Unigrams and bigrams | 66.02 | 20.51 | 31.30 | ✓ | ✓ |
| | Lexicon words | 63.23 | 17.11 | 26.93 | ✗ | ✓ |
| | Unigrams and lexicon words | 70.29 | 20.39 | 31.61 | ✗ | ✓ |
| | Unigrams, bigrams, lexicon words | 68.79 | 24.21 | 35.82 | ✓ | ✓ |
| **Experiment 6: Ensembles of pruned sets approach** | | | | | | |
| SVM | Unigrams | 22.44 | 63.05 | 33.09 | ✓ | ✓ |
| | Unigrams and bigrams | 22.22 | 67.42 | 33.42 | ✓ | ✓ |
| | Lexicon from modified Basilisk | 20.72 | 73.67 | 32.35 | ✓ | ✓ |
| | Unigrams and lexicon words | 23.72 | 85.25 | 37.12 | ✓ | ✓ |
| | Unigrams, bigrams, lexicon words | 16.93 | 71.42 | 27.37 | ✓ | ✓ |
| **Experiment 7: Additional training data with 5-fold cross-validation** | | | | | | |
| SVM | Unigrams | 42.21 | 63.65 | 50.76 | ✓ | ✓ |
| | Unigrams and bigrams | 43.58 | 58.31 | 49.88 | ✓ | ✓ |
| | Lexicon words | 56.06 | 40.41 | 46.97 | ✗ | ✗ |
| | Unigrams and lexicon words | 54.75 | 52.43 | 53.56 | ✗ | ✓ |
| | Unigrams, bigrams, lexicon words | 54.81 | 52.55 | 53.66 | ✗ | ✓ |

## 6.3 Experiments with Semantic Lexicon Approach

Recall that in the semantic lexicon learning approach, we label a report in the test set using the Occurrence Heuristic in combination with the semantic lexicon learned by the modified Basilisk framework described in Section 4.3. Before showing the results of this approach, we first describe how we tune the parameters of the modified Basilisk framework.

### 6.3.1 PARAMETERS

Our modified Basilisk framework has five parameters to tune. The first four are the thresholds resulting from the four frequency-based constraints involving minimum support and maximum generality (see Modification 3 in Section 4.3.3). More specifically, the four "threshold" parameters are (1) the minimum frequency of a word ($Min_W$), (2) the maximum frequency of a word ($Max_W$), (3) the minimum frequency of a pattern ($Min_P$), and (4) the maximum number of words extracted by a pattern ($Max_P$). In addition, recall from Section 4.3.4 that we have three types of patterns (namely, subject-verb/verb-object patterns, bigram patterns for extracting words, and bigram patterns for extracting phrases). Our fifth parameter is the "pattern" parameter, which determines which subset of these three types of patterns to use. Our goal is to tune these five parameters *jointly* on the development set. In other words, we want to find the parameter combination that yields the best F-measure when the Occurrence Heuristic is used to label the reports in the development set. However, to maintain computational tractability, we need to limit the number of values that each parameter can take. Specifically, we limit ourselves to five different combinations of the four "threshold" parameters (see Table 5), and for each such combination, we find which subset of the three types of patterns yields the best F-measure on the development set. Hence the total number of experiments we need to run is 35 (= 7 (the number of (non-empty) subsets from the three types of patterns) × 5 (the number of combinations of the first four parameters)). Our experiment indicates that combination 3 in Table 5, together with the bigram patterns for extracting phrases, yields the best F-measure on the development set, and is therefore chosen to be the best parameter combination involving these five parameters.

The new words and phrases acquired in the first two iterations of modified Basilisk by using this parameter combination are shown in Appendix B. Here we see that no new words are acquired in the first two iterations for eight of the 14 categories. The reasons are that (1) unlike the original Basilisk framework, modified Basilisk employs a common word pool, thus no longer requiring that five words must be added to each category in each bootstrapping iteration; and (2) the application of minimum support to words has led to the filtering of infrequently-extracted words. These two reasons together ensure that the modified Basilisk framework focuses on learning high-precision words for each category.

### 6.3.2 RESULTS

The semantic lexicon learned using the best parameter combination (based on the performance on the development set) is used to label the reports the test set. As we can see from row 1 of Experiment 2 of Table 4, the Modified Basilisk approach achieves a precision of 53.15%, a recall of 47.57% and an F-measure of 50.21%. In comparison to the baseline, this method has a lower precision and a higher recall. The increased recall shows that more reports are covered by the expanded lexicon. However, the learned lexicon also contains some general words that have resulted in a drop in precision. Overall, it has a higher F-measure, which is statistically significantly better than that of the baseline according to both significance tests. This vindicates our premise that learning more words and phrases relevant to the shaping factors will help us identify the shaping factors of more reports.

Table 5: Combinations of the four "threshold" parameters for the modified Basilisk framework.

| Combination | $Min_W$ | $Max_W$ | $Min_P$ | $Max_P$ |
|---|---|---|---|---|
| Combination 1 | 25 | 2500 | 250 | 100 |
| Combination 2 | 25 | 2500 | 100 | 100 |
| Combination 3 | 10 | 2500 | 250 | 100 |
| Combination 4 | 10 | 2500 | 250 | 250 |
| Combination 5 | 10 | 5000 | 250 | 100 |

### 6.3.3 RESULTS USING ORIGINAL BASILISK

To better understand whether our proposed linguistic and algorithmic modifications to the Basilisk framework (see Section 4.3) are indeed beneficial to our cause identification task, we repeated the experiment described above, except that we replaced the lexicon generated using the modified Basilisk framework with one generated using the original Basilisk framework. More specifically, we implemented the original Basilisk framework as described by Thelen and Riloff (2002), but with one minor difference: in the case of the bigram patterns extracting phrases, the word pools described in Section 4.2 were populated with entire phrases instead of only head words. This was done because the seed words list extracted in Section 2.3.2 contains both words and phrases and hence we would like to learn entire phrases.

The only parameter to tune for the original Basilisk framework is the pattern parameter, which, as mentioned above, determines which subset of the three types of patterns to use. Therefore, we construct seven lexicons (corresponding to the seven non-empty subsets of the three types of patterns) using the original Basilisk framework, and determine which lexicon yields the best performance on the development set. Our experiment indicates that the best development result was achieved when only the bigram patterns for extracting phrases were used. Applying the corresponding semantic lexicon in combination with the Occurrence Heuristic to classify reports in the test set, we observe a precision of 49.23%, a recall of 42.78% and an F-measure of 45.78% (see row 2 of the Experiment 2 section of Table 4). This lower precision and higher recall indicates that the lexicon has learned words that are very general (i.e., words that appear in many of the reports and with little discriminative power). The new words and phrases acquired in the first two iterations of original Basilisk are shown in Appendix C. As can be seen, the original Basilisk framework adds a lot of words, but many of them are not relevant to the shaping factors to which they were added, and some are not semantically similar to the seed words for that shaping factor.

Hence, although recall improves by a small amount, precision drops significantly, leading to a precipitation in F-measure. These results suggest that our proposed modifications to the original Basilisk framework are indeed beneficial as far as our cause identification task is concerned.

## 6.4 Experiments with Classification Approach

Recall that in the classification approach to cause identification, we train an SVM classifier for each shaping factor $S_k$ to determine whether a report should be labeled as $S_k$. As desired, this approach allows a report in the test set to potentially receive multiple labels, since the resulting 14 SVM classifiers are applied independently to each report. To investigate the effect of different feature sets on the performance of cause identification, we employ five feature sets in our experiments: (1) unigrams only; (2) unigrams and bigrams; (3) lexicon words only; (4) unigrams and lexicon words; and (5) unigrams, bigrams and lexicon words. The unigrams and bigrams were generated from the reports in the training set by first removing stop-words and ignoring case information, while the semantic lexicon was the one constructed by our modified Basilisk framework. Before showing the results of our supervised and transductive experiments, we first describe the parameters associated with the classification approach.

### 6.4.1 Parameters

For each SVM classifier, we have two parameters to tune. The first parameter is the *percentage* of features to use. Feature selection has been shown to improve performance in text classification tasks (Yang & Pedersen, 1997). As a result, we employ information gain (IG), one of the most effective methods for feature selection according to Yang and Pedersen's experimental results. Since we assume that the words from the semantic lexicon are all relevant to cause identification, we do *not* apply feature selection to the lexicon words. Rather, we apply feature selection only to the unigrams and bigrams. More specifically, if only unigrams are used as features (as in the first of the five feature sets mentioned at the beginning of this subsection), we select the $N\%$ unigrams with the highest IG, where the value of $N$ is tuned using the development set. When both unigrams and bigrams are used as features (as in second and fifth feature sets), we combine the unigrams and bigrams into one feature set and select the $N\%$ unigrams and bigrams with the highest IG, where the value of $N$ is again tuned using the development set. In our experiments, we tested 10 values for $N$: 10, 20, ..., 100.

The second parameter associated with the SVM classifiers is the classification threshold. By default, SVM sets the classification threshold to 0, meaning that every data point with a classification value above 0 is classified as positive, and the rest will be classified as negative. However, since an SVM classifier is trained to optimize classification accuracy, the best classification threshold may not be 0 for our cause identification task, where the goal is to optimize F-measure. As a result, we parameterize the classification threshold, allowing it to take one of 21 values: $-2.0, -1.8, \ldots, 1.8, 2.0$.

As usual, we tune the two parameters described above *jointly* rather than independently. In other words, for each possible value combination of the percentages of features and classification threshold, we compute the F-measure of the classifiers on the development set over all the classes and choose the value pair that yields the maximum F-measure.

To get a better idea of how these two parameters impact performance, we show in Figure 3 how F-measure changes on the development set as we vary the values of the two parameters, from the experiment where the underlying SVM classifiers employ only unigrams as features. As we can see, the best F-measure was achieved by employing the top 50% unigrams and a classification threshold of $-0.8$. Using the default parameter values (no feature selection and a classification threshold of 0) yields a F-measure of approximately 18%. Overall, these results provide suggestive evidence that both parameters can have a large impact on performance.

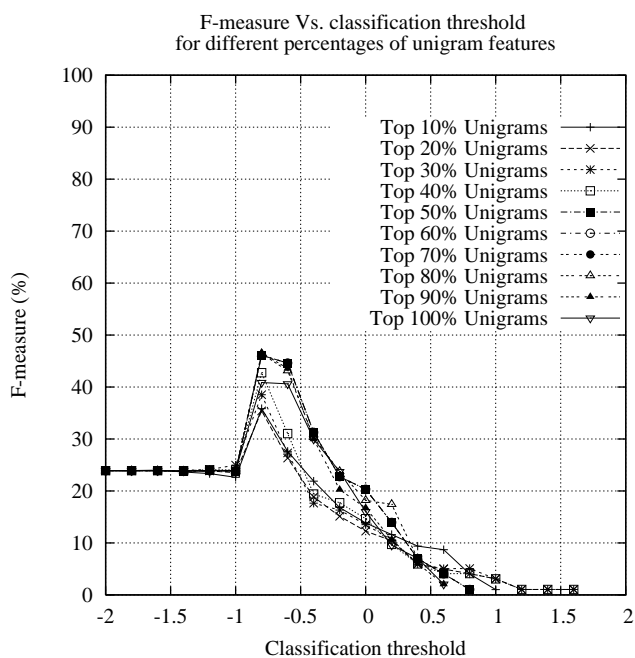

Figure 3: Variation of F-measure with different percentages of unigram features and classification thresholds used for SVM classification.

### 6.4.2 SUPERVISED ONE-VERSUS-ALL CLASSIFIERS: RESULTS AND DISCUSSIONS

Results of the *supervised* One-Versus-All classification approach using the five feature sets described above are shown in the Experiment 3 section of Table 4.[14] As we can see, when feature sets 1 (unigrams only) and 4 (unigrams and lexicon words) are used, we achieve the best results — F-measure scores of 47.46% and 47.45%, respectively. However, even these best results are statistically indistinguishable from the baseline result (according to approximate randomization test), and are significantly worse than the result produced by the modified Basilisk approach (row 1 of Experiment 2) [see Appendix D, which contains statistical significance test results that we obtained by applying stratified approximate randomization test to each pair of experiments in Table 4].

---

14. Recall that in the supervised approach, the SVM classifiers were trained on only the 233 reports in the training set.

In fact, they also indicate that the Occurrence Heuristic has made more effective use of the learned semantic lexicon than the SVM classifiers: the SVM classifiers trained with only the lexicon words as features (row 3 of Experiment 3) produced a significantly worse F-measure score (42.11%) than that of the Occurrence Heuristic (50.21%), due to large drops in both recall and precision. Overall, these results suggest that the supervised approach performs worse than the heuristic-based semantic lexicon approach in this task. We hypothesize that the limited amount of training data available to the SVM learner has contributed to the poor performance of the supervised approach. We will test this hypothesis in Section 6.5

Two additional observations are worth mentioning. First, comparing rows 1 and 4 of Experiment 3, we see that the lexicon words are not useful for cause identification in the presence of unigrams. Second, comparing rows 1 and 2 and then rows 4 and 5 of Experiment 3, we see that using bigrams hurts performance. A likely reason can be attributed to our feature selection method: since we choose the top $N\%$ features, the bigram features significantly outnumber the unigram features, thus potentially diminishing the effect of the latter. One solution to this problem is to employ separate parameters when selecting unigrams and bigrams, but we decided against this choice, as it would lead to an explosion in the size of the parameter space.

### 6.4.3 Transductive One-Versus-All Classifiers: Results and Discussions

To investigate whether it is useful to exploit unlabeled data, we employ transductive SVM to combine labeled and unlabeled data. Essentially, we repeated the experiments in the supervised One-Versus-All classification approach, except that we trained each transductive SVM classifier using both the (labeled) reports in the training set and the (unlabeled) reports in the test set as described in Section 5.2.2. The two parameters — the percentage of features used and the classification threshold — are tuned jointly to maximize F-measure on the development set, as described in the supervised approach, except that the transductive SVMs used in the parameter tuning step are trained using the training set as labeled data and the development set as unlabeled data.

Results of these transductive SVM classifiers are shown in the Experiment 4 section of Table 4. Overall, the transductive results are significantly worse than the corresponding results in Experiment 3. However, the conclusions that we can draw from the transductive results are slightly different from those drawn from the supervised results. First, using bigrams significantly improves performance when the lexicon words are absent (comparing rows 1 and 2 of Experiment 3) but hurts performance when the lexicon words are present (comparing rows 4 and 5). Second, adding lexicon words to the unigram-only feature set (comparing rows 1 and 4) significantly improves performance, suggesting the potential usefulness of the lexicon features. Nevertheless, Experiments 3 and 4 both indicate that (1) using only lexicon words as features are far from adequate, and (2) the best performance is achieved when lexicon words are added to unigrams as features.

### 6.4.4 Results From Additional Supervised Approaches

Next, we present the results from the two additional supervised approaches, namely MetaLabeler and ensembles of pruned sets (Section 5.2.1). The feature sets used by both approaches are the same as those used by the One-Versus-All method. As in the One-Versus-All method, both of these approaches use SVM as the underlying learning algorithm for classifier training.

**MetaLabeler.** The only parameter that needs to be tuned for the MetaLabeler approach is the percentage of features to use ($N$), which was selected based on classification performance (F-measure) on the development set.

Results of the MetaLabeler approach are shown in the Experiment 5 section of Table 4. There are some interesting points about the these results. First, the MetaLabeler method results in much better precision than the other methods. Second, this method shows consistent performance improvement when bigram features are added, as can be seen by comparing the first and second, and fourth and fifth rows of the MetaLabeler results. Third, the inclusion of the lexicon word features are also found to improve performance, as seen by comparing the first and fourth, and second and fifth, rows of the MetaLabeler results. These two observations show that the MetaLabeler approach can properly take advantage of the increasingly richer feature sets used in these experiments, with the best performance occurring when all types of features are used (fifth row). Unfortunately, the approach suffers from poor recall, a fact that prevents it from even matching, let alone surpassing, the F-measure scores of the other methods. Since the method discards the less probable labels when it assigns the labels to the documents, precision is much improved but recall suffers.

**Ensembles of Pruned Set.** Among the parameters of the ensembles of pruned sets approach, the number of classifiers in the ensemble, $M$, and the size of the sample of the training data on which each classifier in the ensemble was trained, were chosen to be the same ones used by Read et al. (2008), namely 10 and 63% respectively. The rest of the parameters of the pruned set approach, namely the minimum cardinality ($b$), the minimum support ($p$), the percentage of features to use ($N$), and the threshold for label assignment ($t$) were selected *jointly* based on classification performance (F-measure) on the development set. The values from which the specific value of $b$ was chosen was 2, 3 and 5. The possible values of $p$ tested in this experiment was 3, 5 and 10. The threshold parameter $t$ was chosen from the values 0.1, 0.2, ..., 1.0, and the percentage of features, $N$ was chosen from the values 10%, 20%, ..., 100%. Thus we had 900 parameter combinations for each feature set, and from these parameter combinations, the combination for which the performance on the development test set was best (in terms of F-measure) was chosen for running the system on the test set.

Results of the pruned set approach are shown in the Experiment 6 section of Table 4. Here, we see the best performance for the combination of unigram and lexicon word features, better than the performance using the unigrams and lexicon words individually. However, performance degraded with the inclusion of bigrams into this combination. Precision is much lower than those of the other methods, which indicates that the selection of the label sets from the training set of only 233 reports may not have been adequate.

### 6.5 Experiments Using Additional Training Data

The results of the above experiments are somewhat surprising: the best-performing supervised classification approach — the One-Versus-All approach — performs significantly worse than the modified Basilisk approach. We hypothesize that its poor performance can be attributed to the scarcity of (labeled) training data. To test this hypothesis, we conducted a set of experiments in which we increased the amount of training data for the One-Versus-All supervised classification approach by applying cross-validation. More specifically, we take the test set of 1000 reports and split it into five disjoint subsets of equal size, $T_1, T_2, \ldots, T_5$. Then, for each $i$ we construct the training set by merging all $T_j$, where $i \neq j$, with the original training set of 233 reports. After that, we train an SVM classifier on this merged training set and test on the set $T_i$. When this is done over all five folds, we compute the F-measure over the entire test set. In other words, the results we report for this set of experiments are not F-measure scores averaged over the five folds. We again experimented with the five set of features used in the supervised experiments in Section 6.4. The two parameters, the percentage of features used and the classification threshold, are tuned in exactly the same way as in the supervised experiments.

Results of this set of experiments are shown in the Experiment 7 section of Table 4. In comparison to the results of Experiment 3, F-measure increases uniformly and significantly. This provides empirical evidence that the performance of the supervised classifiers is limited by the amount of data on which they were trained. With feature sets 4 (unigrams and lexicon words) and 5 (unigrams, bigrams and lexicon words), we achieve the best results — F-measure scores of 53.56% and 53.66% respectively — the difference between which is statistically insignificant. These two results are in turn significantly better than that of modified Basilisk (row 1 of Experiment 2), according to the approximate randomization test. In addition, except for feature set 3 (lexicon words only), results obtained in this experiment are significantly better than that of the baseline, again according to approximate randomization test. Overall, these results suggest the difficulty of the cause identification task: by comparing rows 4 of Experiments 3 and 5, we see that F-measure increases by only about 6% as the number of training reports is increased from 233 to 1033.

A few more points deserve mentioning. As in previous learning-based experiments, using only lexicon words as features yields the worst result in this set of experiments, and combining unigrams and lexicon words still yields one of the best results. Nevertheless, in comparison to Experiment 3, while using bigrams still does not improve performance, it does not *hurt* performance (from a statistical significance point of view). Perhaps more importantly, comparing rows 1 and 4 of Experiment 7, we see that augmenting unigrams with lexicon words yields significantly better performance. This indicates that the lexicon words are indeed useful features for cause identification, but their usefulness may not be revealed when a small labeled training set is used, as seen in Experiment 3. Learning algorithms attempt to learn which features are important or relevant for the given classification task based on the training examples they see, and the more there are training examples, the better they are able to learn the relevance of the features. Our results show a very poignant illustration of this phenomenon: the SVM learner is able to use the lexicon word features effectively only when given a large number of training instances. This can be seen more clearly from the SVM learning curves in Section 6.7.3. This indicates that lexicon words are useful as features only when we have sufficiently large training data. However, lexicon words may still be used effectively in ways other than as linguistic features even if the training set is small, as we can see from the results of Experiment 2, which uses the lexicon words in combination with the Occurrence Heuristic to achieve performances that are statistically significantly better than the baseline.

## 6.6 Error Analysis and Lessons Learned

In order to gain a clearer insight into our cause identification problem and help direct future research, we manually analyzed the errors made by the best-performing system (i.e., the heuristic based approach using the semantic lexicon learned by our modified Basilisk framework) on a randomly chosen 100-report subset of the test set. More specifically, we looked at the *false negatives* (cases in which the annotator labeled a report with a shaping factor but the system did not) and *false positives* (cases in which the system labeled a report with a shaping factor but the annotator did not). For each false negative, we tried to determine why the system failed to correctly label the report, and for each false positive, we tried to determine why the system labeled the report erroneously. Table 6 shows the number of false positives and false negatives along with the reasons for these errors that we discovered in our analysis. The following sections discuss the errors and their reasons in more detail. Note that since a shaping factor may be indicated by more than one keyword in a single report, there can be more than one reason for a false negative (positive) error. Thus the sum of the frequencies of different types of false negative (positive) errors is greater than the total number of false negatives (positives).

Table 6: Error analysis details: different reasons for the false positive and false negative errors.

| False negatives | 58 | Percentage |
|---|---|---|
| Sentence fragments bigger than phrases | 24 | 41.38% |
| Implicit causes that cannot be identified by keywords | 23 | 39.66% |
| Phrases that were not learned | 14 | 24.14% |
| **False positives** | **83** | |
| Keyword was too general | 50 | 60.24% |
| Keyword indicates concept that appears in the report but does not contribute to the incident | 32 | 38.55% |
| Wrongly learned keyword | 6 | 7.23% |
| Keyword was used in a negative context | 3 | 3.61% |
| Keyword was used in a hypothetical context | 1 | 1.20% |

**False negatives.** For each false negative error, we read the report narrative to identify some word, phrase or sentence fragment that may indicate the shaping factor that our system missed. ¿From this analysis, we identified three reasons for the false negatives as follows:

1. **Required sentence fragments larger than phrase.** We identified 24 sentence fragments that are bigger than phrases (i.e., those that consist of two or more phrases). For example, the sentence fragment *having never been to DCA before* consists of 4 phrases: *having never been*, *to*, *DCA* and *before*. Together, they convey the meaning that the reporter was unfamiliar with DCA, but it is not possible to identify a single word or phrase that conveys the same meaning. Since our framework learns only phrases, it was not possible to learn these sentence fragments.

2. **Cause not identifiable by specific words or phrases.** In 21 instances, no specific word, phrase or sentence fragment could be identified that could pinpoint the shaping factors responsible for the incident. For example, a number of reports, including report#566757, describe incidents in which there is a miscommunication between the pilot and the air traffic controller, but that miscommunication must be understood by following their conversation. A human reading the report can easily understand that the pilot is claiming the controller said one thing and the controller is claiming he said something different, but to detect that kind of a scenario, a machine would need to generate a complete model of the discourse that identifies the specific topic of the conversation, the participants, the claims each participant makes about the topic, the fact that the claims are contradictory, and also the fact that the contradiction arises from miscommunication between them. The preprocessed narrative of this report is shown in Appendix E.

3. **Missing phrases.** In 14 cases the necessary phrase was missing from the semantic lexicon learned by our modified Basilisk framework. Out of these 14 phrases, six phrases were too infrequent to be considered by our modified Basilisk framework due to the minimum frequency criterion. For example, the phrase "temperature flux" appears only once in the entire corpus and hence was not considered by our system. Two phrases were verb phrases, which could not have been learned as we focused only on learning noun phrases and adjective phrases. There are four phrases that are not semantically similar to any seed word for their shaping factors. For example, the phrase "garbled transmission" is not semantically similar to any seed word for the shaping factor *Communication Environment*, such as *disturbance*, *static*, *radio discipline*, *congestion* and *noise*. Finally, there are two phrases that should have been learned by the system, but were not learned because at the time they were put into the word pool, other words with higher scores were selected instead.

**False positives.** In the case of false positives, we looked into the report narrative and the keyword that was found in the content to determine why the indication of the shaping factor for the incident described in the report was incorrect. The different reasons that we identified are as follows.

1. **Too general keywords.** We have observed a large number of false positives due to keywords being too *general* (i.e., keywords that have been extracted or learned for a shaping factor but may appear in other phrases that are not related to that shaping factor). For example, the keyword *failure* is a correct indicator of *Resource Deficiency* as it appears in text like "complete electrical failure", "alternator failure", etc., but when it appears in text like "failure to follow Air Traffic Control instructions", it does not indicate *Resource Deficiency* as a shaping factor. We have identified 50 cases that were caused by keywords being too general.

2. **Concept present but not contributing to incident.** Another frequently faced problem is that sometimes the concepts identified by the keywords are present in a report, but they do not act as a shaper for the incident described in the report. For example, in report#324831, the reporter mentions that he was flying *solo*, which is an indication of *Taskload*, but the incident was due to *Physical Environments*, namely snow and foggy weather. The fact that he was flying solo is merely mentioned as a part of his description of the overall situation. The preprocessed version of this report is also given in Appendix E. In total, we observed 32 such cases.

3. **Incorrectly learned words and phrases.** There were six cases in which the semantic lexicon learner learned incorrect words and phrases that were not related to the shaping factors to which they were assigned. For example, the framework incorrectly learned the word *further* for the shaping factor *Resource Deficiency*, and thus a number of reports were mislabeled with *Resource Deficiency*.

4. **Negative context.** There were three cases in which the keyword appeared in a negative context, which is typically signaled by a contextual valence shifter such as "no" and "hardly" (Polanyi & Zaenen, 2006). For example, the keyword *aircraft damage*, an indicator of *Resource Deficiency*, appears in report#569901 as "no apparent aircraft damage", which results in a false positive.

5. **Hypothetical context.** There was one case in which the keyword appeared in a *hypothetical* context in which the reporter conjectures about a possible scenario. The keyword *single pilot*, an indicator of *Taskload*, appeared in report#534432 as "this could happen to a pilot especially if he was single pilot", resulting in a false positive.

**Lessons learned.** Our error analysis provides valuable insight into the nature of the problem as well as hints on how one should proceed in order to improve the performance of the system. By analyzing the most frequent errors, we present the following lessons learned from the analysis. First of all, it is more useful to learn high-precision keywords and phrases than general ones as the largest part of the false positive errors can be attributed to having too general keywords. However, such high-precision keywords and phrases are more likely to have low frequencies, and hence one would have to adapt learning methods to learn useful words and phrases from infrequent ones. Second, one must take into account the fact that relevant portions of the text may be larger than phrases, even going up to clause or sentences. These cannot be identified by learning words or phrases, or N-grams of reasonable size. Thus, more robust methods are needed that can learn useful sentence fragments or useful sentence structures. Finally, there are cases in which one cannot hope to identify using methods that look for keywords, phrases, sentence fragments or even sentence structures, i.e., cases in which the cause of the incident has to be "understood" from the discourse, and cases in which a concept is present in the description and yet plays no part in the incident. Much deeper analysis than simple bag-of-anything models are needed for avoiding these two types of errors, which between themselves represent almost one third of all errors in the analyzed subset. The former needs a method to distinguish relevant sentences from irrelevant ones. For example, Patwardhan and Riloff (2007) discuss a relevant sentence classifier that is trained on a small set of seed patterns and a set of documents marked as relevant and irrelevant that can be useful in this context. The latter problem requires a discourse analysis method that, as discussed earlier, can model the conversations and identify relations correctly. This shows that though it is possible to identify shaping factors from these reports using words and phrases to a certain extent, much deeper natural language techniques are needed to accurately identify the full range of causes.

## 6.7 Additional Analyses

In this section we present the outcomes of a number of additional analyses that we performed on our cause identification task and our approaches to this task. In Section 6.7.1 we study the relative difficulties of classifying the different shaping factors. In Sections 6.7.2 and 6.7.3 we show the learning curves of the semantic lexicon based approach and the learning based approach respectively, i.e., how the performances of these two approaches vary as they are provided with different amounts of training data. Finally in Section 6.7.4 we discuss the outcomes of the experiment conducted to determine if our modifications to the Basilisk framework is useful for learning general semantic categories.

### 6.7.1 PER-CLASS RESULTS

To get an insight into which of the classes are difficult to classify, we perform an analysis of per-class performance of two labeling schemes: the best heuristic-based method (i.e., the Occurrence Heuristic using the lexicon learned by the modified Basilisk framework) [see the first part of Table 7] and the best learning-based method (i.e., the 5-fold SVM classifiers using unigrams, bigrams and the lexicon words as features) [see the second part of Table 7]. In conjunction with Table 1, two classes stand out most prominently as difficult to classify – *Illusion* and *Taskload*. Both of these classes have very little representation in the training, test and development sets, have a very small number of seed words, and result in very poor performance by each of the approaches. The more easily identifiable classes were *Physical Environment*, *Physical Factors*, *Resource Deficiency* and *Preoccupation*, in which both the labeling schemes had F-measures better than 40%. In general these classes had better representation in the training, testing and development sets, and also had a reasonable number of words and phrases in the semantic lexicon. We believe that this difference in characteristics of the classes is a valuable insight that will be helpful in future work.

### 6.7.2 LEXICON LEARNING CURVE

As mentioned in Section 2.3.2, we have used a total of 177 seed words and phrases. At a first glance, this number of seeds may seem large as far as bootstrapping experiments are concerned. However, considering the fact that these 177 seeds are distributed over 14 shaping factors, we only have an average of 12.6 words and phrases per shaping factor. Nevertheless, it would still be interesting to examine how cause identification performance will be affected if we reduce the number of seeds for each shaping factor used by Modified Basilisk in the bootstrapping process. As a result, we ran a set of experiments to measure cause identification performance that uses the semantic lexicon learned by Modified Basilisk when it is given different number of seed words, where the parameters specific to Modified

Table 7: Per-class performance results. The upper table shows the per-class performance of the Occurrence Heuristic using the lexicon learned by the modified Basilisk framework. The lower table shows the per-class performance of 5-fold SVM classifiers using unigrams, bigrams and lexicon words as features.

| Shaping Factor | TP | FN | TN | FP | Precision | Recall | F-measure |
|---|---|---|---|---|---|---|---|
| Attitude | 3 | 27 | 957 | 13 | 18.75% | 10.00% | 13.04% |
| Communication Environment | 9 | 81 | 888 | 22 | 29.03% | 10.00% | 14.88% |
| Duty Cycle | 3 | 23 | 973 | 1 | 75.00% | 11.54% | 20.00% |
| Familiarity | 31 | 19 | 872 | 78 | 28.44% | 62.00% | 38.99% |
| Illusion | 0 | 2 | 996 | 2 | 0.00% | 0.00% | 0.00% |
| Other | 25 | 192 | 766 | 17 | 59.52% | 11.52% | 19.31% |
| Physical Environment | 195 | 70 | 638 | 97 | 66.78% | 73.58% | 70.02% |
| Physical Factors | 22 | 13 | 958 | 7 | 75.86% | 62.86% | 68.75% |
| Preoccupation | 78 | 32 | 822 | 68 | 53.42% | 70.91% | 60.94% |
| Pressure | 14 | 16 | 902 | 68 | 17.07% | 46.67% | 25.00% |
| Proficiency | 40 | 207 | 723 | 30 | 57.14% | 16.19% | 25.24% |
| Resource Deficiency | 360 | 147 | 225 | 268 | 57.32% | 71.01% | 63.44% |
| Taskload | 0 | 29 | 965 | 6 | 0.00% | 0.00% | 0.00% |
| Unexpected | 4 | 6 | 976 | 14 | 22.22% | 40.00% | 28.57% |
| Overall | 784 | 864 | 11661 | 691 | 53.15% | 47.57% | 50.21% |

| Shaping Factor | TP | FN | TN | FP | Precision | Recall | F-measure |
|---|---|---|---|---|---|---|---|
| Attitude | 2 | 28 | 964 | 6 | 25.00% | 6.67% | 10.53% |
| Communication Environment | 20 | 70 | 871 | 39 | 33.90% | 22.22% | 26.85% |
| Duty Cycle | 10 | 16 | 962 | 12 | 45.45% | 38.46% | 41.67% |
| Familiarity | 18 | 32 | 924 | 26 | 40.91% | 36.00% | 38.30% |
| Illusion | 0 | 2 | 998 | 0 | 0.00% | 0.00% | 0.00% |
| Other | 52 | 165 | 685 | 98 | 34.67% | 23.96% | 28.34% |
| Physical Environment | 182 | 83 | 623 | 112 | 61.90% | 68.68% | 65.12% |
| Physical Factors | 20 | 15 | 955 | 10 | 66.67% | 57.14% | 61.54% |
| Preoccupation | 55 | 55 | 848 | 42 | 56.70% | 50.00% | 53.14% |
| Pressure | 6 | 24 | 961 | 9 | 40.00% | 20.00% | 26.67% |
| Proficiency | 102 | 145 | 639 | 114 | 47.22% | 41.30% | 44.06% |
| Resource Deficiency | 399 | 108 | 247 | 246 | 61.86% | 78.70% | 69.27% |
| Taskload | 0 | 29 | 971 | 0 | 0.00% | 0.00% | 0.00% |
| Unexpected | 0 | 10 | 990 | 0 | 0.00% | 0.00% | 0.00% |
| Overall | 866 | 782 | 11638 | 714 | 54.81% | 52.55% | 53.66% |

Basilisk are set as described in Section 6.3.1. More specifically, we chose the top 3, 4, 5, 6, 7, 10, 15 and 20 seed words and phrases for each shaping factor (in terms of the frequency in the entire corpus), and ran the modified Basilisk framework for ten iterations using the aforementioned parameters.

Note, however, that not all shaping factors have the same number of manually selected seed words and phrases. For example, *Illusion*, *Taskload* and *Unexpected* have 1, 2 and 3 seed words and phrases respectively, whereas *Resource Deficiency* and *Physical Environment* have 47 and 45 respectively (see the last column of Table 1). Hence, for those experiments where the number of seeds used for each shaping factor exceeds the number of manually selected seeds for a shaper, all the manually selected seeds were used. For example, since *Unexpected* has only three manually selected seeds, all of them were used in the experiments in which at least three seeds are used for each shaping factor.

The Occurrence Heuristic was then used with the lexicons thus generated to evaluate their performance on the test set. The resulting learning curve, in terms of F-measure on the test set of 1000 reports, is shown in Figure 4. In addition, since the baseline to which we compare the performance is based on the seed words, the baseline learning curve corresponding to each reduced seed words set is also shown. As expected, increasing the number of seed words monotonically improves the F-measure. However, the improvement over the baseline is particularly small when fewer than seven seed words are used, and the highest improvement is observed for seven seed words and phrases. ¿From then on, adding more seeds improves the overall performance, but the improvement over the baseline slowly diminishes.

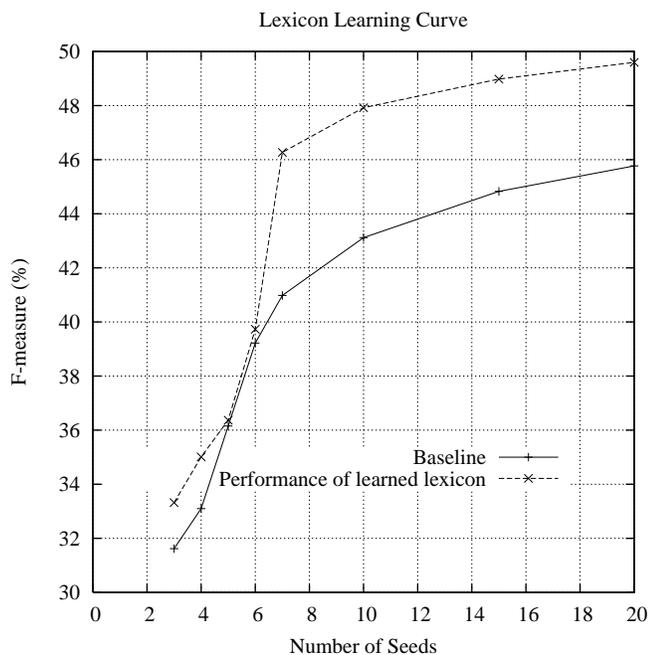

Figure 4: Variation of F-measure with different number of seeds words per category.

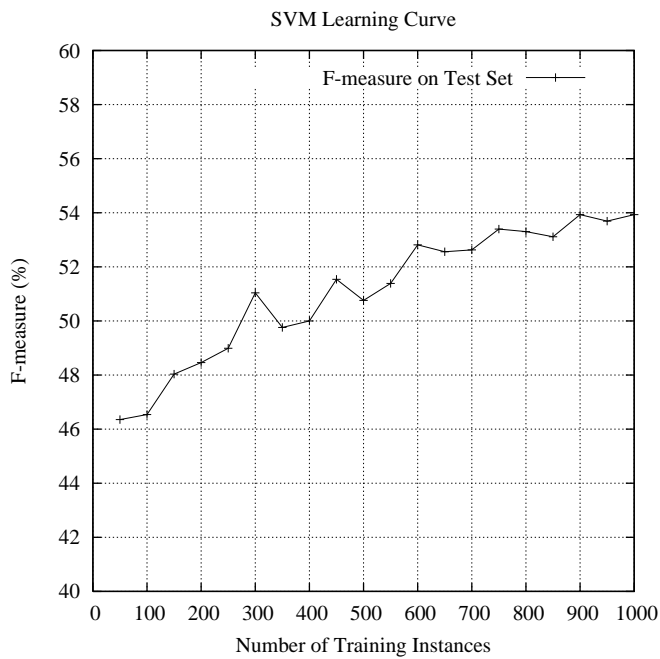

Figure 5: Variation of F-measure with different number of training reports.

### 6.7.3 SVM Learning Curve

As discussed in Section 6.5, we hypothesize that the failure of the SVM classifiers to perform better than the baseline is due to the scarcity of the training instances available to the learner. One may argue that SVM has been shown to work well for small datasets. So, a natural question is: how much smaller will the training set be before we can see a statistically significant drop in cause identification performance? To answer this question, we plot a learning curve for the One-Versus-All classification approach, using as features a combination of unigrams, bigrams, and lexicon word features in a five-fold cross validation setting, which is the setting that yields the best performance in Table 4. Specifically, we generated random subsets of the training sets of sizes 50, 100, ..., 1000 instances. Parameters, namely the percentage of features and the classification threshold, were chosen in the same way as the original experiment as described in Section 6.5, and the F-measure was evaluated on the entire test set. The curve is shown in Figure 5, where each data point is computed by averaging the results over five independent runs. As we can see, there is a general trend of performance improvement with the increase in the number of training instances. In addition, when trained on only 50% of the training set, we see that the cause identification system started to perform statistically significantly worse than the system that was trained on all of the available instances according to the stratified approximate randomization test.

### 6.7.4 GENERAL USEFULNESS OF OUR MODIFICATIONS TO BASILISK

In order to test whether the modifications we made to the Basilisk framework are useful to lexicon learning in general, we added two general categories to the shaping factors in the bootstrapping experiments, namely *People* and *Equipment*. These two categories were added because, firstly, words and phrases added to these categories would be easy to verify (i.e., whether they are words or phrases representing people or equipment), and secondly, they are similar to the original context in which the Basilisk framework was originally evaluated (i.e., learning words in the categories BUILDING, EVENT, HUMAN, LOCATION, TIME, and WEAPON from terrorism reports). These two additional categories were added to the seed lexicon described in Section 2.3.2, which was then bootstrapped by running Original Basilisk and Modified Basilisk separately for ten iterations, with the parameters specific to these Basilisk frameworks set in the same way as described in Section 6.3.1. The seed words for these two categories were selected in the same manner as done by Thelen and Riloff (2002), i.e., the phrases in the corpus were sorted by frequency and the five most frequent phrases belonging to these categories were manually identified. Below are the seed words used in the two categories:

- **People:** Captain, controller, First Officer, RPTR, passenger

- **Equipment:** aircraft, airplane, Collision Avoidance System II, engine, Auto-Pilot

In order to verify which of the words and phrases learned by the two frameworks correctly belong to the assigned category, the first author and a computer science graduate student not affiliated with this research went over the generated lexicons. Appendices F and G show the lexicons generated by Original Basilisk and Modified Basilisk respectively. To facilitate analysis, we divide the words and phrases in each generated lexicon into three categories: (1) those that are determined as correct by both human judges; (2) those that are determined as correct by only one judge; and (3) those that are determined as incorrect by both judges.

For the lexicon generated by Original Basilisk, we find that in the category *People*, 29 of the 50 words and phrases were determined as correct by both judges, and 6 were determined by exactly one of the judges as correct; in the category *Equipment*, 6 of the 50 words and phrases were correct according to both judges, and 22 were correct according to exactly one of the judges. On the other hand, for the lexicon generated by Modified Basilisk, we find that in the category *People*, 44 of the 50 words and phrases were determined as correct by both judges, and 3 were determined by exactly one of the judges as correct; in the category *Equipment*, 34 of the 50 words and phrases were correct according to both judges, and 9 were correct according to exactly one of the judges. This comparison clearly shows that the modifications that we made to the Basilisk framework are not specific to this particular task; rather, these modifications have improved the lexicon building performance in general.

## 6.8 Summary of Conclusions

We end this section by providing a summary of the major conclusions we draw from the experiments.

- Our heuristic approach to cause identification, which labels a report using the Occurrence Heuristic in combination with the words and phrases automatically acquired using our Modified Basilisk framework, surpasses the performance of the baseline system, which applies the Occurrence Heuristic in combination with the seed words and phrases manually identified from the training documents. The difference in F-measure between these two systems is statistically significant according to both McNemar's test and the stratified approximate randomization test. This suggests that the words and phrases in the semantic lexicon learned via Modified Basilisk are *relevant* and *effective* for cause identification.

- Adding the learned lexicon words to an N-gram-based feature set for training SVM classifiers is beneficial for cause identification only if the training set is sufficiently large, as exhibited by the statistically significant increase in F-measure. This provides suggestive evidence that the words and phrases in the semantic lexicon learned via Modified Basilisk are *relevant* and *useful* features for cause identification.

- When used in combination with the Occurrence Heuristic, the semantic lexicon learned by our Modified Basilisk framework offers significantly better performance for the cause identification task than the one obtained using the original Basilisk framework. Additional experiments reveal that Modified Basilisk is not only useful for cause identification, but it also offers performance superior to Original Basilisk when bootstrapping general semantic categories such as *People* and *Equipment*.

- Among the three multi-class multi-labeled text classification approaches we experimented with, One-Versus-All works significantly better than MetaLabeler and Pruned Sets for cause identification. Transductive learning, when used in combination with the One-Versus-All approach, significantly hurts performance, suggesting that unlabeled data cannot be profitably exploited given the fairly small amount of labeled data.

- Our best system achieves an F-measure of around 53.7%, which indicates that cause identification is a difficult task, and that there is a lot of room for improvement. To provide directions for future research, we performed an analysis of the errors made by the best-performing system. In particular, we found that performance is currently limited in part by several factors. First, there are a number of cases in which the relevant text indicating the responsible shaping factor may be larger than phrases. Second, indicators for a shaping factor may be mentioned in a report without influencing the incident described in the report. Finally, there are some cases in which the shaping factors cannot be identified by simply looking at the words, phrases or even sentence fragments – much deeper analysis is required in these cases.

- Increasing the number of seed words and phrases employed by Modified Basilisk improves cause identification performance, but the marginal performance improvement for each added seed diminishes with successive additions. In other words, these results seem to suggest that using more seed words will be unlikely to improve much over the current performance; rather it would be more promising to start with a small number of high frequency seeds and improve upon the bootstrapping process.

- The learning curve plotted for the One-Versus-All classification approach shows that cause identification performance increases with the number of training instances. In particular, when trained on only 50% of the training set, we see that the resulting cause identification system performs statistically significantly worse than the one trained on all of the available instances.

Overall, while our approaches rely on automatically learned lexicon words and phrases that are not adequate for cause identification, they are *relevant* for the task. As mentioned previously, their use is motivated by the labor-intensive procedure that the NASA researchers employed in manually identifying seed words and phrases for each shaping factor (Posse et al., 2005). Our work represents one of the first attempts to tackle this cause identification task, and we believe that the use of simple features is a good starting point and establishes a baseline against which future studies on this problem can be compared.

The main take home message from this research is that though it is possible to solve the problem that we set out to solve, namely automated cause identification, by learning relevant keywords or sentence fragments and other suitable bag-of-words models, there remains a significant portion of the data that remains unlabeled or mislabeled through these methods. To match the performance level achieved in other topical text classification tasks, much deeper linguistic analysis like relevant sentence detection and discourse analysis methods like identifying disagreements, disputes and hostile attitudes will be needed. This lesson should be the cornerstone of further research in this area.

## 7. Related Work

In this section, we describe some other works related to our research. In particular, our discussion focuses on causal analysis as well as approaches to semantic lexicon construction and text classification, and is organized as follows. First, we discuss causal analysis as it has appeared in different fields. Second, we discuss the different semantic lexicon learning algorithms. Third, we discuss works that deal with extraction pattern learning. Fourth, we describe different algorithms for unsupervised word clustering and thesaurus building. Finally, we include a discussion of related work on multi-class multi-labeled text classification.

**Causal analysis.** Major research on causality has been performed mainly in the fields of philosophy and psychology. In the field of philosophy, seminal works in causality have been conducted by Hume (1739, 1748), who provides one of the most influential definitions of cause as "*an object followed by another, and where all the objects, similar to the first, are followed by objects similar to the second. Or, in other words, where, if the first object had not been, the second never existed.*" This has been the basis of later, much stronger definitions of causation (e.g., Lewis, 1973; Ganeri, Noordhof, & Ramachandran, 1996). Notable investigations on causation in the field of psychology include those by Cheng (1997), who defines causation in terms of the *probabilistic contrast model*; Griffiths and Tenenbaum (2005), who discuss learning about cause and effect relationships using causal graphical models; and Halpern and Pearl (2005), who provide explanations of causality by means of structural equations governing random variables representing events. Although these works provide important background and definitions contributing to the understanding of causality, in order to identify causes from naturally written text we must turn to NLP.

In the field of NLP, there is little work on cause identification similar to our problem. Research on causality focuses mainly on identifying causal relations between two sentence components. For instance, Girju (2003) describes a method for automatically discovering lexico-semantic patterns that refer to causation. In particular, she focuses on the explicit intra-sentential pattern $\langle NP_1 \text{ verb } NP_2 \rangle$, where *verb* is a simple causative. She also shows how these patterns can be used to improve the performance of a system for answering cause-effect type questions. Khoo et al. (2000) use graphical pattern matching to identify causal relations from medical article abstracts. They use hand-crafted patterns that are matched against the parse trees of the sentences. The subtrees of the parse tree that match the patterns are extracted as causes or effects. Similarly, Garcia (1997) uses hand-crafted extraction patterns to identify causal relations from sentences in the French language. The limitation of these approaches is that they focus on identifying causal relations from the same sentence, whereas our reports are multi-sentence discourses.

Grishman and Ksiezyk (1990) use domain modeling, discourse analysis and causal inference to find cause-effect relations between events leading up to equipment malfunctions from short equipment failure reports. More specifically, they first apply syntactic analysis to produce parse trees for the sentences in the reports using an augmented context-free grammar. Then they apply semantic analysis to map (1) verbs and syntactic relations into domain-specific predicates and relations and (2) noun phrases into references to components in the domain model. Finally, they apply discourse analysis to these predicates to construct a *time-graph* showing the temporal and causal relationships between the elementary facts. The temporal relations are derived from text structures and words (e.g., "when", "then", etc.) and the order of appearance in text, but the causal relations are determined by querying a simulation model of the equipment that is built using domain knowledge. Specifically, each possible causal link is posed as a query to the model to test if the relation holds. Overall, their method relies heavily on the domain model of the equipment being studied, and their research focuses on only one specific piece of equipment.

NASA's own research on identifying causes of incidents from the report narratives have been performed by Posse et al. (2005), who describe a specific experiment in which they brought together experts to manually analyze the report narratives and identify words, phrases and expressions related to each of the shaping factors, as mentioned earlier. Later work by Ferryman et al. (2006) take these manually extracted expressions as ground truth and compare the anomalies described in the reports to the shaping factors derived from applying these expressions to the same reports. Specifically, they do not attempt to learn these expressions automatically; rather, they focus on finding possible correlations between the shaping factors and the anomalies.

**Algorithms for semantic lexicon learning.** A number of semantic lexicon learning algorithms follow an iterative bootstrapping approach, starting from a small number of semantically labeled seed words. Roark and Charniak (1998) propose a method of constructing semantic lexicons based on co-occurrence statistics of nouns in conjunctions, lists and appositives. They start with a small seed nouns list, and iteratively add similar words to that list. The word similarity is measured by the ratio of how many times the word occur together with a seed word to the number of times the word appear in the corpus. After construction, they rank the words by a log-likelihood statistic (Dunning, 1993). However, due to the general brevity of the reports, such co-occurrences and lists are rather few in our corpus, and it is more useful for us to use context-based similarities like Thelen and Riloff (2002). They describe their *Basilisk* framework for learning semantic lexicon using extraction patterns as features. The apply a weakly supervised bootstrapping approach in which they start from a small manually constructed seed lexicon and iteratively add semantically similar words to it. This, described in more detail in Section 4.2, has been the basis for our lexicon learning approach.

A number of improvements to the Basilisk framework, and more generally to bootstrapping approaches, have been proposed. In the Basilisk framework, the number of iterations is a parameter that has to be chosen arbitrarily. Rather than making an arbitrary choice, Yangarber (2003) proposes a method for detecting termination of unsupervised semantic pattern learning processes. The method requires that the documents must be labeled as relevant or irrelevant. Since such information is not available for our corpus, it is not useful for us. Curran, Murphy, and Scholz (2007) suggest an improvement over traditional bootstrapping methods by discarding words and contexts that appear to be related to more than one category, in order to minimize semantic drift and enforce mutual exclusion. On the other hand, we handle such cases by comparing the conditional probabilities for the different categories to which such words can belong. Zhang, Zhou, Huang, and Wu (2008) present bootstrapping with the graph mutual reinforcement-based bootstrapping (GMR) (Hassan, Hassan, & Emam, 2006), a modification of the Basilisk method. Similar to us, they explore using N-grams to capture context, but they use a different set of pattern and word scoring formulas. For learning multiple categories simultaneously, they introduce a scoring system based on entropy of a pattern. They report better results than Basilisk on the MUC-4 dataset (see Sundheim, 1992).

Among non-bootstrapping approaches, Ando (2004) presents a new method of constructing semantic lexicons from an unannotated corpus using a set of semantic classes and a set of seed words and phrases for each semantic class. She uses spectral analysis to improve the feature vectors by projecting the useful portions of the vectors into a subspace and removing the "harmful" portions of the vectors. The resultant feature vectors are then used by a centroid-based classifier using cosine similarity measure to label the words. Avancini, Lavelli, Sebastiani, and Zanoli (2006) take a classification approach to semantic lexicon construction. They cast the problem as a term (meaning both words and phrases) categorization task (dual of the document categorization task), and similar to the bag-of-word model, they represent the terms as bag-of-documents. They use a variation of the adaptive boosting algorithm, *AdaBoost.MH*$^{KR}$, which is trained on a small seed lexicon and then used to classify the noun terms in the corpus to zero, one or more semantic categories.

**Algorithms for learning extraction patterns.** Our approach to semantic lexicon construction uses extraction patterns as features, and here we present some methods that aim to improve the extraction pattern collection process. Riloff (1996) describes the *AutoSlog-TS* system that learns extraction patterns from untagged text. However, it needs a *pre-classified* corpus that have the text classified as relevant or irrelevant; as we mentioned earlier, we do not have access to such information. Phillips and Riloff (2007) present a method of boot-strapping algorithm to learn *role-identifying nouns*, which are then used to learn important extraction patterns, and also role-identifying expressions. However, their focus is mainly on identifying the roles of words in events.

Patwardhan and Riloff (2007) provide another extraction pattern learning approach using relevant regions. They require the documents to be pre-classified into relevant and irrelevant documents. Using a small set of seed patterns, they classify the sentences in these documents into relevant and irrelevant sentences. Then "semantically appropriate" extraction patterns are learned using a *semantic affinity* metric and separated into primary and secondary patterns. This approach is also not directly usable to us due to the unavailability of documents pre-classified into relevant and irrelevant categories.

Recently, the Internet has increasingly been used in natural language research. Patwardhan and Riloff (2006) use the *AutoSlog-TS* system (Riloff, 1996) to learn domain specific extraction patterns by processing documents retrieved by querying the web with selected domain-specific words. Using the web is an interesting and promising enhancement and, as mentioned in Section 8, we intend to extend our work using the Google corpus (Brants & Franz, 2006).

**Algorithms for thesaurus building and unsupervised word clustering.**    Another area of research that is very closely related to semantic lexicon learning is *thesaurus building*. Building a thesaurus requires discovering groups of semantically similar words, though it stops short of assigning semantic class labels to the words. Thus it shares the problem of measuring semantic similarity and grouping similar words with the semantic lexicon building task. Here we discuss several approaches to the thesaurus building task.

Clustering has been used extensively in thesaurus building, mostly because of its unsupervised nature and ability to handle large volumes of data. The seminal work in this direction has been by Pereira, Tishby, and Lee (1993), who present an unsupervised method for soft clustering of words using distributions of the words in different contexts. This approach generates overlapping word clusters, grouping words based on the contexts that they appear in. Baker and McCallum (1998) use Pereira et al.'s distributional clustering technique to perform feature space reduction for supervised classification with naïve Bayes by using the clusters as features. Lin and Pantel (2001) present their approach of generating a collection of sets of semantically similar words, or *concepts*, using their clustering method, *UNICON*, with dependency relations as features. Pantel and Lin (2002) present another clustering approach, *clustering by committee*, using contextual features with point wise mutual information as feature values, that they compare as better than Lin and Pantel's results. Rohwer and Freitag (2004) present a clustering-based automatic thesaurus building process from an unannotated corpus. They propose an *information theoretic co-clustering* algorithm that groups together highly frequent words into clusters of similar part-of-speech category. Then they pursue an additional process, *lexicon optimization*, to grow the lexicon by assigning the less frequent words to their most likely clusters.

Among non-cluster-based methods, Davidov and Rappoport (2006) present an unsupervised method of discovering groups of words that have similar meanings. They achieve this by (1) identifying *high frequency words* and *content words*, (2) identifying *symmetric* lexical relationship patterns, and (3) applying a graph clique-set algorithm to generate word categories from co-occurrence information of the content words in the symmetric patterns.

Concentrating on the performance issues that plague attempts to build thesaurus from a large corpus, Rychlý and Kilgarriff (2007) present two methods of improving performance of general context-based thesaurus building algorithms. The first method is to compare only those word pairs that have some context in common. The second method is to use the heuristic of removing those contexts that are too general (i.e., contexts that have more than a certain number of distinct words). In our research, we have adopted the second method (see Section 4.3). They also applied a partitioned sequential approach to the construction process. Though thesaurus building does not usually require an annotated corpus or a set of seed words and phrases, it is not directly applicable to the task of growing a semantic lexicon where we have to learn words in specific semantic categories. This is because the method has no control over which words are being learned and which classes the discovered word groups belong to. It may be possible to adapt this method to that of semantic lexicon growing by classifying the word groups into the semantic classes by using the seed words and phrases. However, the method has to be extended to extract noun and adjective phrases.

**Algorithms for multi-class multi-labeled text classification.** As mentioned previously, cause identification, when cast as a text classification problem, is a multi-class multi-labeled text classification problem, since there are 14 shaping factors in total and each document may be labeled with more than one shaping factor. There are several popular approaches to solving a multi-class multi-labeled text classification problem. The first, and one of the approaches followed in this research, is to independently train a binary classifier for each class, and apply each classifier on a test instance in isolation. In our case, the underlying learner is Support Vector Machines (Joachims, 1998). Godbole and Sarawagi (2004) suggest a number of improvements to this scheme, namely, including class labels suggested by a preliminary set of classifiers as features, removing negative examples too close to the classification hyperplane, and selectively removing some classes from the one-versus-others classifications scheme. Another notable method, followed by Tsoumakas and Vlahavas (2007) and Read et al. (2008), is to treat each unique set of labels as a new label, thus converting the problem to a multi-class single-labeled one. Their works differ from each other in the construction of the new labels. The former, called *RAndom k-LabELsets*, or *RAKEL*, builds an ensemble of classifiers by randomly sampling label sets of size $k$; whereas the latter adopts the method of filtering observed label sets by minimum support. Tang et al. (2009), on the other hand, take a different approach: they train one classifier that predicts the number of labels that a test instance would have, and then choose that many labels for that instance based on output of another classifier that ranks the labels by likelihood for that instance. All these works use SVM as their underlying learner. In addition, all these approaches make the assumption that the classes are correlated to a high degree. However, an analysis of our dataset does not present evidence of such a strong correlation. Of the 140 documents with multiple labels in the test set, there are 68 unique label sets, of which only seven have a frequency of at least five. Thus increasing the number of labels would only aggravate an already imbalanced class distribution.

Among other approaches, we mention two systems that use probabilistic generative models. McCallum and Nigam (1999) propose a system that starts with a small set of keywords and unlabeled documents, and learns a naïve Bayes classifier in a bootstrapping process from the keyword-induced labels by using hierarchical shrinkage and expectation maximization on a held-out data set. Ueda and Saito (2002) present another generative model called *Parametric Mixture Models*, which treats multi-labeled text as a parametric mixture of words relevant to each label. Their work is closely related to *Latent Dirichlet Allocation* (Blei, Ng, & Jordan, 2003). The generative models usually assume that a document related to a particular topic would have a high frequency of words related to that topic. In our research, the documents are mostly devoted to description of the event that occurred, and the cause of that event is only mentioned briefly. This makes generative models less suitable for the task at hand as generative models would more likely generate models related to the events and not the causes. A more comprehensive review of different approaches to multi-class multi-label text classification can be found in the work of Tsoumakas and Katakis (2007).

## 8. Conclusions

We have investigated two approaches to the cause identification task, the goal of which is to understand why an aviation safety incident happened via the identification of the causes, or shaping factors, that are responsible for the incident. Both approaches exploit information provided by a semantic lexicon, which is automatically constructed via Thelen and Riloff's (2002) Basilisk framework augmented with our three algorithmic modifications (namely, the use of a probabilistic similarity measure, the use of a common word pool, and the enforcement of minimum support and maximum generality constraints for words and extraction patterns) and one linguistic modification (the use of N-gram-based extraction patterns). The heuristic-based approach labels a report by employing the Occurrence Heuristic, which simply looks for the words and phrases acquired during the semantic lexicon learning process in the report. The learning-based approach labels a report by employing inductive and transductive support vector machines to learn models from reports labeled with shaping factors. Our experimental results indicate that the heuristic-based approach and the supervised learning approach (when given sufficient training data) both significantly outperform our baseline, which, motivated by NASA's work, labels a report simply by using the Occurrence Heuristic in combination with a set of manually-identified seed words and phrases. More importantly, results of the heuristic-based approach indicate that our modifications to the original Basilisk framework are beneficial as far as cause identification is concerned, and results of the learning-based approach indicates the usefulness of the lexicon words when they are used in combination with unigrams as features for training an SVM classifier. Overall, what we set out to prove was that it is possible to automate the cause identification task by manually analyzing a small number of reports and using the information thus generated to train machine learning methods to identify the shaping factors of the rest of the reports. Our experiments have been able to prove the feasibility of this approach, and also the usefulness of learning a semantic lexicon and using the words in it as features. Nevertheless, our best system achieves an F-measure of around 53.7%, which indicates that cause identification is a difficult task, and that there is a lot of room for improvement. In particular, our analysis of the errors made by the best system on 100 randomly chosen test documents provides valuable insights into the task as well as directions for future research.

From our experience from this current research, we intend to extend our work in the following directions. First and foremost, we plan to extend our approach to handle text fragments bigger than phrases. Second, in order to improve the quality of labeling, we propose to work on improving the lexicon learning performance further by using different semantic similarity measures. For instance, we would like to study the performance of the semantic similarities and weighting functions suggested by Curran and Moens (2002) in our context. Third, we plan to use more thoroughly normalized text for better parsing and tagging, as well as relevant region information (Ko, Park, & Seo, 2004; Patwardhan & Riloff, 2007). Fourth, we propose to augment the semantic lexicon, specifically by using the Google N-grams corpus (Brants & Franz, 2006) to extract frequent N-gram patterns for words. Fifth, we propose to explore other more recent lexicon construction methods like unsupervised word clustering (Pantel & Ravichandran, 2004), spectral analysis, mutual exclusion bootstrapping, co-clustering and exploiting symmetric patterns. Finally, in order to handle the shaping factors that are difficult to identify from words occurring in the reports, we propose to employ much deeper analysis of the text at the semantic level. We have also taken the step of making our annotated incident reports publicly available, and we hope that we can stimulate research on this under-investigated problem in the NLP community.

## Acknowledgments

The authors would like to thank the anonymous reviewers who provided us with comments that were invaluable in improving the quality of the paper. This research was supported in part by NASA grant NNX08AC35A. Any opinions, findings, conclusions or recommendations expressed in this paper are those of the authors and do not necessarily reflect the views or official policies, either expressed or implied, of NASA.

## Appendix A: Seed Words

Below are the seed words manually extracted from the 233 reports in the training set (see Section 2.3.2 for details).

| Shaping Factor | Seed words |
|---|---|
| Attitude | get HOMEITIS, attitude, inattentiveness, get THEREITIS, complacency, overconfidence, sarcastic, inattention |
| Communication Environment | disturbance, static, radio discipline, congestion, noise |
| Duty Cycle | 11 hour duty day, inadequate rest, last of 4 legs, heavy flying, reduced rest, all-night flight, 12 hour day, red eye, ten leg day, all night |
| Familiarity | familiarization, not familiar, new, first departure, unfamiliar, unfamiliarity, very familiar, low time, first landing |
| Illusion | bright lights |
| Other | noise abatement policy, disoriented, confused, medical emergency, economic considerations, disorientation, drunk passenger, confusion |
| Physical Environment | cold, clouds, dark, setting sun, sun glare, obscured, visibility, hazy stratus, birds, fog bank, solid overcast, snow, weather, rime, gust, low weather, surface winds, jet blast, lightning, sea gulls, high ceilings, hot, tailwind, chop, very dark, sea gull, winds, scattered, high tailwinds, extremely dark, too bright, icing, turbulence, RPTED wind, terrain, bird strike, crosswind, thunderstorm, glare, reduced visibility, high flying birds, fog, severe winter weather, cloud, ice |
| Physical Factors | very tired, HYPOXIA, tiredness, tired, fatigued, disorientation, fatigue, no rest |
| Preoccupation | distracted, preoccupied, mental lapse, busy, DISTRS, distraction, attention, inattention, absorbed |
| Pressure | hurry, running late, pressure, low on fuel, fuel considerations, behind schedule, late, peer pressure, under pressure, rushing |
| Proficiency | mistakes, mistaken, new hire, inexperience, forgotten, less than 100 hours, newly rated, training, recent pilot, inadvertently, bad turn, MISINTERPED |
| Resource Deficiency | loose connection, erratic, blown, overheated, bang, collapse, no idea, unavailable, placarded, crack, Out Of Service, damage, smoke, inoperative, failure, leak, deferred items, communication failure, loss, unreliable, FDRS problem, bump, shaking, master caution, inadequately lighted, unreadable, disconnected, malfunction, shudder, absence, hazard, inaccurate, UNFLAGGED, fire, broken, fluctuations, compressor stall, deferral, unusable, wrong, intermittently, warning, discrepancies, faulty, deferred, intermittent, missing |
| Taskload | single pilot, solo |
| Unexpected | unexpected, suddenly, UNFORECAST |

## Appendix B: Sample Lexicon Words Learned by Modified Basilisk

Below are the semantic lexicon words learned by the modified Basilisk framework in the first two iterations.

| Shaping Factor | New words |
|---|---|
| Attitude | |
| Communication Environment | |
| Duty Cycle | |
| Familiarity | aligned, fairly new, more familiar |
| Illusion | |
| Other | initial confusion, minimum fuel emergency, misunderstanding, weather emergency |
| Physical Environment | TRWS, conflict message, cumulonimbus, large cells, numerous thunderstorms, occasionally severe, thunderstorm cells, weather buildups, weather cell, weather en route |
| Physical Factors | first factor |
| Preoccupation | adequate attention, as much attention, close attention, close enough attention, crew attention, enough attention, much attention, proper attention, strict attention, very close attention |
| Pressure | |
| Proficiency | |
| Resource Deficiency | different, amiss, awry, obviously wrong, resulting loss, seriously wrong, slight loss, temporary loss, terribly wrong, very wrong |
| Taskload | |
| Unexpected | |

## Appendix C: Sample Lexicon Words Learned by Original Basilisk

Below are the semantic lexicon words learned by the original Basilisk framework in the first two iterations.

| Shaping Factor | New words |
|---|---|
| Attitude | Air Traffic Control security, aileron yoke displacement, anomalous VFR Omni-Directional Radio Range information, assured TFR avoidance, betrayal, concern and urgency, forgetting air carrier X, magnified problem, operational complacency, unseen and unknown turbulence |
| Communication Environment | 9001 noise, BNA runway 31 approach plate, LIGHTSPEED 20K noise, Non- noise, OVERSPD bell, active noise, clearance delivery transmission, left engine stall, static and Emergency Locator Transmitter, stuck trim or elevator movement |
| Duty Cycle | 10 plus 16 layover, 2 different time frames, 3 'back-to-back ' continuous duty trips, 4 hour break, 69 minutes, 8 hour 15 minute flight time day, Pacific Daylight Time departure, TPA flight, XC15 departure, scheduled 2- leg continuous duty |
| Familiarity | S partial unfamiliarity, 'S perceived familiarity, Command familiarity, Command unfamiliarity, blue panel indication light, dispatch work desks, generally unfamiliar, inexperience and unfamiliarity, new everyday, past experience and familiarity |
| Illusion | 1 1/2 Nautical Mile SSW, 1/2-1/4 point, 10 ' off end, 50 feet side, Elmendorf required use, about 1/2 mile downwind, airspace E, foxtrot intersection, lateral boundaries, mile right |
| Other | misinformation, Flight Management System/heading anomalies, confusion/conflict, disoriented and confused, intense panic, micro sleep, miss numerous times, mistake inconvenience, note closure problem, start terror |
| Physical Environment | MHT class Celsius, STRATO-cumulus, Thur morning, clouds underneath, compacted snow and ice, fair weather cumulus, next morning weather, puffy cumulus clouds, thin scattered clouds, well developed cumulus clouds |
| Physical Factors | HYPOXIA/carbon monoxide, Minimum Equipment List 24-32-02, basically tired, cardiac distress, indicating system problem, internal bleeding, interrupted fuel flow, oncoming seizure, stress overload, upper respiratory problems |
| Preoccupation | Captain and First Officer attention, Terminal Radar Approach Control Facility distraction, close enough attention, consequently my attention, good enough attention, lip service, mind or attention, much mind, real attention, real close attention |
| |  |

| Shaping Factor | New words |
|---|---|
| Pressure | Minimum Equipment List pressure, consistent answer, elevator pressure, intense pressure, part # Coordinated Universal Time, repercussion, right engine pressure, significant pressure, slow gear, wheel pressure |
| Proficiency | & P school, CL65 ground school, basic flight training, hard lesson, initial and annual proficiency training, occurrence and strive, rating training, several military flying clubs, situation event, time limited simulator sessions |
| Resource Deficiency | Air Traffic Control loss, altitude deviation/loss, apparently inoperative, either inoperative, even a reexamining, intermittent or inoperative, known traffic conflict or loss, observed loss, recently a Los, thankfully accurate |
| Taskload | 16500 # turboprop, A320 type aircraft, AVIAT husky A1 two place tail DRAGGER aircraft, Cessna 402 type aircraft, Cessna model 421 type aircraft, L1011-250, LGA-MHT flight, McDonnell Douglas MD11, more solo cross country FLTS, solo cross country privileges |
| Unexpected | significant, 1 jolt, approximately 5-10 sec, choppy and aircraft, consistently moderate, contributing workload factor, industry issue, just as severe, rapid and immediate, real cushion |

## Appendix D: Additional Stratified Approximate Randomization Tests

To ascertain the statistical significance of the difference between the F-measure scores of the different report labeling methods, we performed the stratified approximate randomization test with 9,999 shuffles between all pairs of the results of Experiments 1 through 5 in Table 4. The table below shows if the method in the column is statistically significantly better than the method in the row at the level of $p < 0.05$. As before, statistical significance and insignificance are denoted by a ✓ and an X, respectively.

| Method[a] | SW | MLW | OLW | SVM-U | SVM-UB | SVM-L | SVM-UL | SVM-UBL | SVMT-U | SVMT-UB | SVMT-L | SVMT-UL | SVMT-UBL | SVM5-U | SVM5-UB | SVM5-L | SVM5-UL | SVM5-UBL |
|---|---|---|---|---|---|---|---|---|---|---|---|---|---|---|---|---|---|---|
| SW | - | ✓ | X | X | X | X | X | X | X | X | X | X | X | ✓ | ✓ | X | ✓ | ✓ |
| MLW | X | - | X | X | X | X | X | X | X | X | X | X | X | X | X | X | ✓ | ✓ |
| OLW | X | ✓ | - | X | X | X | X | X | X | X | X | X | X | ✓ | ✓ | X | ✓ | ✓ |
| SVM-U | X | ✓ | X | - | X | X | X | X | X | X | X | X | X | ✓ | ✓ | X | ✓ | ✓ |
| SVM-UB | ✓ | ✓ | X | ✓ | - | X | ✓ | X | X | X | X | X | X | ✓ | ✓ | X | ✓ | ✓ |
| SVM-L | ✓ | ✓ | ✓ | ✓ | ✓ | - | ✓ | X | X | X | X | X | X | ✓ | ✓ | ✓ | ✓ | ✓ |
| SVM-UL | X | ✓ | X | X | X | X | - | X | X | X | X | X | X | ✓ | ✓ | X | ✓ | ✓ |
| SVM-UBL | ✓ | ✓ | ✓ | ✓ | X | X | ✓ | - | X | X | X | X | X | ✓ | ✓ | ✓ | ✓ | ✓ |
| SVMT-U | ✓ | ✓ | ✓ | ✓ | ✓ | ✓ | ✓ | ✓ | - | ✓ | ✓ | ✓ | ✓ | ✓ | ✓ | ✓ | ✓ | ✓ |
| SVMT-UB | ✓ | ✓ | ✓ | ✓ | ✓ | X | ✓ | X | X | - | X | ✓ | X | ✓ | ✓ | ✓ | ✓ | ✓ |
| SVMT-L | ✓ | ✓ | ✓ | ✓ | ✓ | ✓ | ✓ | ✓ | X | ✓ | - | ✓ | ✓ | ✓ | ✓ | ✓ | ✓ | ✓ |
| SVMT-UL | ✓ | ✓ | X | ✓ | X | X | ✓ | X | X | X | X | - | X | ✓ | ✓ | ✓ | ✓ | ✓ |
| SVMT-UBL | ✓ | ✓ | ✓ | ✓ | ✓ | X | ✓ | X | X | X | X | X | - | ✓ | ✓ | ✓ | ✓ | ✓ |
| SVM5-U | X | X | X | X | X | X | X | X | X | X | X | X | X | - | X | X | ✓ | ✓ |
| SVM5-UB | X | X | X | X | X | X | X | X | X | X | X | X | X | X | - | X | ✓ | ✓ |
| SVM5-L | X | ✓ | X | X | X | X | X | X | X | X | X | X | X | ✓ | ✓ | - | ✓ | ✓ |
| SVM5-UL | X | X | X | X | X | X | X | X | X | X | X | X | X | X | X | X | - | X |
| SVM5-UBL | X | X | X | X | X | X | X | X | X | X | X | X | X | X | X | X | X | - |

a. Legend: SW = Occurrence Heuristic using seed words, MLW = Occurrence Heuristic using modified Basilisk lexicon, OLW = Occurrence Heuristic using original Basilisk lexicon, SVM-U = SVM using unigrams, SVM-UB = SVM using unigrams and bigrams, SVM-L = SVM using lexicon words, SVM-UL = SVM using unigrams and lexicon words, SVM-UBL = SVM using unigrams, bigrams and lexicon words, SVMT-U = transductive SVM using unigrams, SVMT-UB = transductive SVM using unigrams and bigrams, SVMT-L = transductive SVM using lexicon words, SVMT-UL = transductive SVM using unigrams and lexicons, SVMT-UBL = transductive SVM using unigrams, bigrams and lexicon words, SVM5-U = 5-fold SVM using unigrams, SVM5-UB = 5-fold SVM using unigrams and bigrams, SVM5-L = 5-fold SVM using lexicon words, SVM5-UL = 5-fold SVM using unigrams and lexicon words, SVM5-UBL = 5-fold SVM using unigrams, bigrams and lexicon words.

## Appendix E: Sample Preprocessed Reports

### Report ACN#324831

RETURNING to waukegan regional airport from practice area located between 5-20 mile W of the airport; flying solo as a student pilot; at about 3000 feet Mean Sea Level Visual Flight Rules. cloud area about 5 mile W of the airport obscured view ahead so I reduced altitude to proceed Visual Flight Rules and returned to 3000 feet after passing the thin cloud line. the area to the n; containing fix references for airport location; was shrouded in clouds and in fog at ground level. the same was true of the lake michigan shoreline to the E. the ground was also substantially snow covered. although the airspace over the airport was undoubtedly clear; as was the practice area; orientation to the field was lost to me. I climbed to 4500 feet to increase the overview; without benefit. returning to 3000 feet; I flew to what I believed to be n of the airfield to landfall the airport. I must have been to the S; however; and proceeding S I flew into ord class B airspace. coinciding to being lost; I contacted waukegan tower; not then realizing that I had flown as Federal Aviation Regulation S as I had. I was directed to contact ord approach on the frequency given; and beginning with ord approach was vectored back to waukegan airport; frequency changed to tower control and blessedly cleared to land. the time lost was between 1 hour 15 minutes and 1 1/2 hours.

### Report ACN#566757

THE following event occurred while REPOSITIONING; by taxi; from the W side to the S side of isp airport. I initially contacted longitude island tower asking permission to REPOS from the W side to the OPS Base Operations Office of the tower (the S side). the controller replied 'start taxi via taxiway W up to but hold short of runway 6.' I read back the instructions stating to start taxi via taxiway W holding short of runway 6. as I was taxiing; there was an aircraft on taxiway W holding short of runway 6; performing a run-up. the controller asked if I was able to get around the aircraft. I replied that I was able. the controller then said 'use caution taxiing around the aircraft and cross runway 6.' as I was taxiing across runway 6; I noticed an aircraft on short final for runway 6. I was clear of the runway before the aircraft touched down. the controller then came on the frequency and said 'you were instructed to hold short of runway 6.' I replied 'you cleared me across runway 6.' the controller said 'call the tower when you park.' I replied 'roger; I will call you when I park.' I called and talked to the controller a few minutes later and he said 'you were instructed to hold short of runway 6.' again I told him that he had cleared me across the runway. I feel that pilots and controllers need to listen to each other and decipher what is said before acting on it.

## Appendix F: Lexicon Learned by Original Basilisk for Categories *People* and *Equipment*

The following table shows the words and phrases learned by the original Basilisk framework for the categories *people* and *equipment* (see Section 6.7.4).

| Category | New words |
|----------|-----------|
| People | **Agreed by *both* judges as *correct*:** ABQ tower procedure specialist, ACN 126721 reporter, AFSFO, AVP tower specialist, Air Route Traffic Control Center specialist, Air Traffic Control facility reps, BDR tower specialist, BHM control, BUF field operations officer, CAE tower specialist, Chicago quality control, DFW maintenance manager, Flight Service Station dispatcher, SFOLM the Captain, SII program manager, Stearman pilot, TLH supervisor, bur local controller, casino manager, cos Air Traffic Control chief, flight test engineers, local balloon repairman, outbound Captain and First Officer, repair facility and pilot, shift boss, spokesperson, station management individual, technician desk, tower supervisor/manager **Identified by *one* judge as *correct*:** Flight Standards District Office ORL, again maintenance supervisor, approach controller verbatim, freighter aircraft and approach, him and tower, passenger and fatigue **Agreed by *both* judges as *incorrect*:** ACN 670635, ACN 682482, AT6 aircraft, B737-300/500 SRM, EMB service manual, Non-air carrier aircraft, RPTR ACN 518698, RPTR ACN 601074, RPTR ACN 658075, RPTR ACN 664336, RPTR ACN 676343, RPTR ACN 88920, cabin or company, other aircraft center, reliable research resources |
| Equipment | **Agreed by *both* judges as *correct*:** Collision Avoidance System II 10 Distance Measuring Equipment screen, Collision Avoidance System II B737-200, Collision Avoidance System II EHSI, Collision Avoidance System II IVSI display, Collision Avoidance System II RR, Collision Avoidance System II Missed Approach Point page, Collision Avoidance System II RR, |
| |  |

| Category | New words |
| --- | --- |
|  | **Identified by *one* judge as *correct*:** Collision Avoidance System II VSI overlay, 'Resolution Advisory ' stopped and aircraft, Collision Avoidance System II 'stop climb ' alert, Collision Avoidance System II 'traffic ; ' then 'climb ' advisory, Collision Avoidance System II 'traffic ; traffic ' aural warning, Collision Avoidance System II 3 mile circle, Collision Avoidance System II 5 mile scale, Collision Avoidance System II 6 mile scale, Collision Avoidance System II POPUP traffic, Collision Avoidance System II Resolution Advisory alerts, Collision Avoidance System II Resolution Advisory climb priority, Collision Avoidance System II Resolution Advisory climb warning, Collision Avoidance System II Resolution Advisory signals, Collision Avoidance System II Resolution Advisory zone, Collision Avoidance System II Resolution Advisory/Traffic Advisory alert, Collision Avoidance System II Resolution Advisory/Traffic Advisory alerts/advisories, Collision Avoidance System II Resolution Advisory/altitude deviation, Collision Avoidance System II Traffic Advisory and Resolution Advisory alerts, Collision Avoidance System II WINDSHEAR warning, Collision Avoidance System II advisory alert, Collision Avoidance System II advisory alert and warning, Collision Avoidance System II warning and aircraft **Agreed by *both* judges as *incorrect*:** Collision Avoidance System II 10 O'clock and 2 1/2 to 3 mile, Collision Avoidance System II Resolution Advisory 'climb ' command, Collision Avoidance System II Resolution Advisory area, Collision Avoidance System II Resolution Advisory climb or descent, Collision Avoidance System II Resolution Advisory data tag, Collision Avoidance System II Resolution Advisory descent, Collision Avoidance System II Resolution Advisory green band target, Collision Avoidance System II Resolution Advisory increase climb, Collision Avoidance System II Resolution Advisory maneuvering, Collision Avoidance System II Resolution Advisory messages, Collision Avoidance System II Resolution Advisory recovery procedure, Collision Avoidance System II Resolution Advisory requirement, Collision Avoidance System II Resolution Advisory requiring climb, Collision Avoidance System II Resolution Advisory resolution, Collision Avoidance System II Traffic Advisory notification, Collision Avoidance System II Traffic Advisory/Resolution Advisory aircraft, Collision Avoidance System II Traffic Advisory/Resolution Advisory event, Collision Avoidance System II action requirements, Collision Avoidance System II advice, Collision Avoidance System II advisories and instructions, Collision Avoidance System II caution, Collision Avoidance System II quit |

## Appendix G: Lexicon Learned by Modified Basilisk for Categories *People* and *Equipment*

The following table shows the words and phrases learned by the modified Basilisk framework for the categories *people* and *equipment* (see Section 6.7.4).

| Category | New words |
|---|---|
| People | **Agreed by *both* judges as *correct*:** ' First Officer, ' my First Officer, ; First Officer, CP, Captain, Captain RPTR, Captain trainee, Co-Captain, Co-pilot, First Officer, First Officer # 2, Initial Operating Experience Captain, PAXS, Pilot Flying and First Officer, Potomac controller, RPTING Captain, RPTING First Officer, RPTING pilot, RPTR Captain, RPTR pilot, S/O, ZOA supervisor, air carrier Y pilot, aircraft X pilot, aircraft commander, all the passenger, analyst, and First Officer, baron pilot, biplane pilot, controller, facility person, first observer, flight attendant # 3, flight attendants and passenger, flying Captain, forward observer, passenger, passenger and crew, passenger and flight attendants, right seat pilot, second observer, sic, specialist, student Captain, supervisor/Controller, tower Controller, tower operator, training pilot <br> **Identified by *one* judge as *correct*:** RPTR, gate and passenger, so First Officer <br> **Agreed by *both* judges as *incorrect*:** departure and departure, neither the Captain, which CLRLY |
| Equipment | **Agreed by *both* judges as *correct*:** # 1 Auto-Pilot, # 2 Auto-Pilot, 3 AUTOPLTS, AUTOFLT, AUTOTHROTTLE, AUTOTHROTTLE and Auto-Pilot, AUTOTHROTTLES, AUTOTHROTTLES and Auto-Pilot, AUTOTHRUST, Auto-Pilot # 1, Auto-Pilot # 2, Auto-Pilot B, Auto-Pilot and AUTOTHRUST, Auto-Pilot and PMs, Auto-Pilot and throttles, Auto-Pilot/AUTOTHROTTLES, Cessna 180, Collision Avoidance System II system, ENGS # 2 and # 3, aircraft ABCD, aircraft Auto-Pilot, aircraft engine, allowed aircraft, automatic pilot, automatic throttle, automatic throttles, autopilot, center Auto-Pilot, craft, emergency engine, left Auto-Pilot, left hand engine, parked plane, right Auto-Pilot <br> **Identified by *one* judge as *correct*:** problem engine, maintenance and aircraft, later aircraft, aircraft and aircraft, Collision Avoidance System II alert, # 1 Constant Speed Drive, Auto-Pilot and AUTOTHROTTLES, WDB 2, perf <br> **Agreed by *both* judges as *incorrect*:** aircraft beginning, aircraft parallel, normal and aircraft, person or property, persons or property, so aircraft, time aircraft |